\documentclass[12pt,american]{article}

\usepackage[T1]{fontenc}
\usepackage[utf8]{inputenc}
\usepackage[lf]{ebgaramond}

\usepackage[letterpaper]{geometry}
\usepackage{float}
\usepackage{url}
\usepackage{amsmath}
\usepackage{amssymb}
\usepackage{graphicx}
\usepackage{bbm}
\usepackage{textcomp}
\DeclareUnicodeCharacter{2212}{\textminus}
\usepackage{booktabs}
\usepackage{indentfirst} 
\usepackage{setspace}
\makeatletter
\@ifundefined{date}{}{\date{}}
\newcommand{\betweensize}{\@setfontsize\betweensize{9.5}{9.5}} 
\makeatother

\usepackage{titletoc}
\usepackage[ page,header]{appendix}

\usepackage{subcaption}
\usepackage{multirow}

\usepackage{tabularx}
\usepackage{framed}
\usepackage{longtable}
\usepackage[flushleft]{threeparttable}
\usepackage{color}
\usepackage{booktabs,caption}
\definecolor{Crimson}{RGB}{220, 20, 60}
\definecolor{Maroon}{RGB}{195, 33, 72}
\usepackage{hyperref}  

\usepackage{chronosys}
\usepackage{hhline}
\usepackage{xcolor}
\usepackage{colortbl}
\usepackage{float}

\usepackage{pdflscape}
\usepackage{txfonts}
\hypersetup{
  colorlinks=true,
  citecolor={blue},
urlcolor={blue},
linkcolor ={blue}}

\usepackage{array}

\usepackage{bibentry}

\usepackage{tikz}
\usepackage{tikz-qtree,tikz-qtree-compat}
\usetikzlibrary{arrows}
\usetikzlibrary{shapes.multipart}

\usepackage{subcaption}
\usepackage{dsfont}
\usepackage{CJKutf8}
\usepackage[overlap, CJK]{ruby}

\usepackage{listings}
\newenvironment{SChinese}{%
\CJKfamily{gkai}%
\CJKtilde
\CJKnospace}{}

\makeatother

\usepackage{babel}
\usepackage{csquotes} 
\usepackage[
  backend=biber,
  style=chicago-authordate,
  doi=false,
  eprint=false,
  isbn=false,
  maxcitenames=2,
  mincitenames=1,
  maxbibnames=99
]{biblatex}
\DeclareFieldFormat{yearhyperref}{\bibhyperref{#1}}

\DeclareCiteCommand{\textcite}
  {\usebibmacro{prenote}}
  {%
    \printnames{labelname}%
    \space%
    (\printtext[yearhyperref]{\printfield{year}})%
  }
  {\multicitedelim}
  {\usebibmacro{postnote}}

\DeclareCiteCommand{\parencite}
  {\usebibmacro{prenote}\bibopenparen}
  {%
    \printnames{labelname}%
    \space%
    \printtext[yearhyperref]{\printfield{year}}%
  }
  {%
    \multicitedelim
  }
  {%
    \usebibmacro{postnote}%
    \bibcloseparen
  }

\graphicspath{{figures/}}

\begin{document}

\pagenumbering{gobble}

\title{Agentic Framework for Political Biography Extraction}
\author{Yifei Zhu\thanks{Ph.D candidate, Department of Politics and Public Administration, The University of Hong Kong. Email: \url{frankyifei@connect.hku.hk}.} \and Songpo Yang\thanks{Boya Postdoctoral Fellow, School of International Studies, Peking University. Email: \url{yangsp21@mails.tsinghua.edu.cn}.} \and Jiangnan Zhu\thanks{Associate Professor, Department of Politics and Public Administration, The University of Hong Kong. Email: \url{zhujn@hku.hk}.} \and Junyan Jiang\thanks{Assistant Professor, Department of Political Science, Columbia University. Email: \url{jj3160@columbia.edu}.}}

\maketitle
\singlespacing
\begin{abstract}
    The production of large-scale political datasets typically demands extracting structured facts from vast piles of unstructured documents or web sources, a task that traditionally relies on expensive human experts and remains prohibitively difficult to automate at scale. In this paper, we leverage Large Language Models (LLMs) to automate the extraction of multi-dimensional elite biographies, addressing a long-standing bottleneck in political science research. We propose a two-stage ``Synthesis-Coding'' framework for complex extraction task: an upstream synthesis stage that uses recursive agentic LLMs to search, filter, and curate biography from heterogeneous web sources, followed by a downstream coding stage that maps curated biography into structured dataframes. We validate this framework through three primary results. First, we demonstrate that, when given curated contexts, LLM coders match or outperform human experts in extraction accuracy. Second, we show that in web environments, the agentic system synthesizes more information from web resources than human collective intelligence (Wikipedia). Finally, we diagnosed that directly coding from long and multi-language corpora introduces bias that the synthesis stage can alleviate by curating evidence into signal-dense representations. By comprehensive evaluation, We provide a generalizable, scalable framework for building transparent and expansible large scale database in political science.

\end{abstract}

\begin{center}
\end{center}
\doublespacing
\newpage
\setcounter{page}{1} 
\renewcommand{\thepage}{\arabic{page}}
\newcolumntype{x}[1]{>{\centering\let\newline\\\arraybackslash\hspace{0pt}}p{#1}}

\begin{refsection}

\section{Introduction}
The empirical revolution in political science has been fueled by the mass analysis of digitized political records \parencite{grimmer2013text, gentzkow2019text}. Disclosed government documents, digitized news reports, and crawled web pages enable large-scale research using political facts\footnote{By ``political facts,'' we refer to verifiable, descriptive attributes of political actors and institutions, including office holding, educational backgrounds, career paths, and institutional affiliations, that can be reliably documented and cross-checked across sources.}, facilitating theory building on representation, state capacity, and regime durability \parencite{binderkrantz2024closed,fisman2020social,jiang2018making,jiang2020friends,nyrup2025paths}. Yet transforming unstructured document stacks into structured, analyzable datasets remains prohibitively labor-intensive. The core bottleneck is fact extraction: researchers must gather evidence from sources, extract specific information, and structure verifiable information into data suitable for downstream analysis, a process that currently requires extensive trained manual labor. Scaling political data production beyond its current bottleneck demands automated solutions that can replicate, and potentially exceed, the validity of human coding while dramatically reducing labor costs.

This paper develops and evaluates automated LLM-based solutions for the extraction of political facts from unstructured documents at scale, focusing on one consequential class of political facts: structured elite biography. Elite behaviour, shaped by backgrounds, incentives, and networks, systematically influences policy making, public opinion, and regime stability \parencite{putnam1976comparative,svolik2012politics,king2013censorship,alexiadou2015ideologues,reuter2019elite,jiang2018making,woldense2024elite}. Structured elite biographies are analytically rich for downstream analysis, but computationally prohibitive under traditional manual methods. Manually constructing a political biography requires gathering related sources, distinguishing the right sources, extracting biographical facts, including entities, events, and relations from unstructured text into temporally organized lists.\footnote{While we focus on elite biographies, the challenges identified here, such as information dispersion, temporal inconsistency, source conflict, and extraction from unstructured text, apply broadly to other ``narrative'' political data, such as tracking policy evolution, coding event data from news reports, or reconstructing negotiation processes from diplomatic cables.} In a landmark study, \textcite{nyrup2025paths} mobilized over 30 research assistants across three years to manually assemble the ``Paths to Power'' (PtP) dataset on cabinet members worldwide from diverse web resources.\footnote{The sheer scale of such efforts is often understated. Coding a comprehensive cabinet dataset typically requires tens of thousands of RA hours. For instance, the WhoGov dataset \parencite{nyrup2020governs} took nearly a decade of intermittent work to finalize.} While these achievements, together with many other similar efforts \parencite{armstrong2024financial,back2021war,jiang2018making,lee2022breaking,raleigh2022elite,vittori2023technocratic}, have proven indispensable for answering theoretical questions across diverse political contexts, the need for manual extraction imposes severe limitations, such as discontinued or out of date datasets, costly coverage expansion and new variable addition.

Recent advances in Large Language Models (LLMs) offer a potential path toward scalable political text processing \parencite{benoit2025using,ornstein2025train, gilardi2023chatgpt, palmer2024using}. However, existing political-science applications focus on \textit{classification} tasks, where output is one  label from a predefined finite set \parencite{benoit2025using,halterman2024codebook, ziems2024can}. The harder question is whether LLM-based systems can perform valid extraction tasks, with multiple fields codebooks and undefined output spaces. Extraction in political science also features long, noisy document collections and even web resources, where naive zero shot or few shot fails to address \parencite{liu2024lost}.

To address automation of extraction tasks without sacrificing validity, we propose and evaluate an agentic framework for political facts extraction from web sources. The agentic framework consists of two stages: an upstream \textit{synthesis} stage that utilizes agentic recursive LLM calls to search and refine evidence from web sources, and a downstream \textit{coding} stage that maps that refined evidence into structured facts. We evaluate this framework against human extraction using a validated ground truth dataset of political elites biographies from China, the United States, and a comparative sample of OECD countries. We first validate the \textit{coding} ability of LLMs when the input is human curated Wikipedia biography, showing that LLM coders can match and exceed human coding quality using curated short corpora. We then prove that the agentic workflow outperforms human collective synthesis (Wikipedia) in producing curated biography corpora for global political elites. Finally, utilizing two coding corpora containing same information but different composition, we test the generalizability of the two stage (synthesis and coding) framework for extraction task, showing that long and multi-language corpora can introduce coding quality degradation, which proper synthesis can mitigate.

Our contribution is fourfold. First, we formalize political fact extraction as distinct from classification. Whereas classification typically assigns texts to a fixed set of labels, extraction requires identifying and structuring facts from open-domain sources and assembling them into coherent career histories. Second, we propose a Synthesis-then-Coding framework that treats information synthesis as a prerequisite for valid extraction and clarifies why skipping synthesis (e.g., naive context-window stuffing) induces a quantity--quality trade-off that degrades performance. Third, we develop and open-source a scalable agentic package that operationalizes synthesis through iterative, tool-using retrieval and refinement, and we show that it can outperform human collective synthesis (Wikipedia) in producing high-signal corpora for global political elites. Fourth, we apply this framework to generate a large cross-national dataset of political elite biographies, lowering the barrier to producing and maintaining high-quality data in information-poor environments and providing a generalizable template for extracting structured narratives from unstructured text.

\section{The Challenge of Extracting Political Biographies at Scale}

To understand why automatic solutions are necessary, it is useful to begin with the type of political facts that pose the greatest challenges for large-scale data production. Among the many categories of political facts, elite biographical information (e.g., who political elites are, where they come from, and how they advance through institutions), constitutes a particularly demanding case. Such data form the backbone of comparative political research. Granular information on educational backgrounds, career trajectories, and kinship networks has been central to theories of political representation \parencite{carnes2024white, lee2022breaking}, authoritarian power-sharing \parencite{svolik2012politics, raleigh2022elite}, and technocratic governance \parencite{lin2020rise, vittori2023technocratic}. In the study of Chinese politics, for example, detailed career histories have enabled scholars to uncover the logic of factional patronage \parencite{jiang2018making,shih2012getting} and to assess the regime’s claims of meritocratic selection \parencite{liu2024meritocracy}.\footnote{While our discussion focuses on national-level elites (e.g., cabinet ministers) due to data availability, the theoretical importance of biographical data extends to local officials, bureaucrats, and party cadres \parencite{landry2008decentralized}. The scalability constraints we identify are arguably even more severe for these lower-tier populations, where $N$ is larger and data are noisier.}

Despite their importance, elite biographical data remain exceptionally difficult to produce at scale. While digitization has expanded access to political texts,\footnote{We distinguish between \textit{digitization} (converting physical records into digital text) and \textit{extraction} (converting unstructured text into structured databases). The latter remains the primary bottleneck for narrative political data.} extraction continues to rely overwhelmingly on manual coding by experts or research assistants. Crowdsourcing platforms offer an alternative labor model, but they have generally proven unsuitable for complex elite data extraction, which requires substantial domain knowledge to resolve ambiguities in names, titles, and political affiliations \parencite{benoit2016crowd}. These constraints translate into extraordinary labor requirements in practice. The ``Paths to Power'' dataset \parencite{nyrup2025paths}, covering cabinet ministers in 141 countries over 55 years, required more than five years of coordinated work by over 30 research assistants. The LEAD dataset \parencite{ellis2015introducing} assigned multiple coders to each leader and still took three years to complete. Other prominent efforts, including \textcite{funke2023populist} and \textcite{braun2010banking}, likewise required years of intensive manual verification. As summarized in Table~\ref{tab:elite_datasets}, high-quality elite datasets typically mobilize large teams over extended periods and depend on sustained institutional funding from agencies such as the European Research Council or U.S. National Science Foundation \parencite{alexiadou2022cabinet,ellis2015introducing}.

\begin{table}[htbp]
\centering
\caption{Available Datasets on Political Elites Since 2007$^{*}$}
\label{tab:elite_datasets}
\footnotesize
\resizebox{\textwidth}{!}{%
\begin{tabular}{@{}lccccl@{}}
\toprule
Dataset & N & Years$^{\dagger}$ & Region & Variables$^{\ddagger}$ & Production Details \\
& Countries & & & (D/C/I/P) & \\
\midrule
\multicolumn{6}{l}{\textit{National Leaders}} \\

\textcite{baturo2016cursus} & -- & 1960--2010 & Global & \checkmark/\checkmark/--/\checkmark & 2 RAs, 2009--14 \\
\textcite{baturo2022what} & -- & 1950--2017 & Global & --/\checkmark/--/-- & 1 RA \\
\textcite{baturo2023incumbent} & 132 & 1918--2019 & Global & --/--/--/\checkmark & Built on 11 datasets \\

\textcite{bomprezzi2025wedded} & 177 & 1989--2018 & Global & \checkmark/--/--/\checkmark & 27 RAs, 1.6M entities \\

\textcite{deluca2018ethnic} & 140 & 1992--2013 & Global & \checkmark/--/--/-- & Augmented Archigos \\
\textcite{dreher2009impact} & 72 & 1970--2002 & Global & \checkmark/\checkmark/--/-- & -- \\

\textcite{ellis2015introducing} & 188 & 1875--2004 & Global & \checkmark/\checkmark/--/-- & 2 RAs/leader, 3 yrs \\
\textcite{eschenauer2023coup} & -- & 1950--2020 & Global & --/--/--/\checkmark & -- \\

\textcite{fearon2007ethnic} & 161 & 1945--1999 & Global & \checkmark/--/--/-- & -- \\
\textcite{funke2023populist} & 60 & 1900--2020 & Global & --/--/\checkmark/-- & 9 RAs, 20k+ pages \\

\textcite{gerring2019rules} & 162 & 2010--2013 & Global & \checkmark/\checkmark/--/-- & -- \\
\textcite{goemans2009introducing} & 188 & 1875--2015 & Global & --/--/--/\checkmark & -- \\

\textcite{herre2023identifying} & 182 & 1945--2020 & Global & --/--/\checkmark/-- & 15 RAs \\

\textcite{licht2022introducing} & -- & 1960--2015 & Global & --/--/--/\checkmark & 5 grad students \\

\textcite{mattes2016measuring} & 169 & 1919--2018 & Global & --/--/--/\checkmark & 9 RAs+Experts \\

\textcite{yu2020rich} & 177 & 1946--2011 & Global & \checkmark/\checkmark/--/-- & -- \\

\midrule
\multicolumn{6}{l}{\textit{Sub-National \& Ministerial-Level Elites}} \\

\textcite{alexiadou2015ideologues} & 18 & 1945--2013 & OECD & --/\checkmark/--/-- & -- \\
\textcite{alexiadou2022cabinet} & 18 & 1945--2015 & OECD & \checkmark/\checkmark/--/-- & Multiple coders, 6 yrs \\
\textcite{alexiadou2022technocratic} & 13 & 1980--2014 & W. Europe & --/\checkmark/--/-- & Multiple experts \\

\textcite{armstrong2024financial} & 191 & 1972--2017 & Global & \checkmark/\checkmark/--/-- & 6 RAs \\

\textcite{back2021war} & 13 & 1789--2021 & Great Powers & \checkmark/\checkmark/--/-- & -- \\

\textcite{braun2010banking} & 154 & 1996--2005 & Global & --/--/--/-- & 72,769 names checked \\

\textcite{carozzi2016sending} & 1 & 1994--2006 & Italy & \checkmark/--/--/-- & -- \\

\textcite{fuchs2018development} & 23 & 1967--2012 & OECD & \checkmark/\checkmark/\checkmark/-- & 10 RAs \\

\textcite{hallerberg2012educational}  & 27 & 1973--2010 & OECD & \checkmark/--/--/-- & 6 RAs \\

\textcite{jiang2018making} & 1 & 1997--2015 & China & \checkmark/\checkmark/--/\checkmark & 20+ RAs \\

\textcite{lee2022breaking} & 4 & 1983--2017 & Asia & \checkmark/\checkmark/--/-- & -- \\

\textcite{nyrup2020governs} & 177 & 1966--2023 & Global & --/--/\checkmark/-- & 9+ coders \\
\textcite{nyrup2025paths} & 141 & 1966--2021 & Global & \checkmark/\checkmark/--/-- & 30+ RAs, multi-year \\

\textcite{raleigh2022elite} & 23 & 1996--2017 & Africa & --/--/--/-- & -- \\

\textcite{ricart2021colonial} & 16 & 1960--2010 & Africa & \checkmark/--/--/-- & -- \\

\textcite{vittori2023technocratic} & 31 & 2000--2020 & EU+4 & \checkmark/\checkmark/--/-- & Country experts \\

\bottomrule
\end{tabular}
}
\vspace{0.5em}
\begin{minipage}{\textwidth}
\footnotesize
$^{*}$This list is illustrative rather than exhaustive. We prioritize datasets that (1) focus on individual-level attributes of political elites, (2) have been widely cited in top political science journals, and (3) involve substantial manual coding efforts. Datasets focused solely on voting records (e.g., roll-call data) are excluded as they represent a different class of ``atomic'' facts.

$^{\dagger}$\textit{Years} reflect the temporal coverage of the most recent available version. 
The end year indicates when the dataset was last updated, which may postdate the cited 
foundational paper (e.g., Archigos 4.1, initially published by \textcite{goemans2009introducing}, 
was subsequently updated to cover leaders through 2015). Most datasets have not been 
updated for several years, with many remaining frozen a decade or more behind current events.

$^{\ddagger}$\textit{Variables}: D = Demographics (education, ethnicity, birthplace, family background); 
C = Career (pre-office occupation/political experience); 
I = Ideology/party affiliation; 
P = Power dynamics (entry/exit manner, tenure, transitions). 
Production scale indicates the reported labor intensity of manual data collection; 
-- = Not reported or insufficient detail.
\end{minipage}
\end{table}

Even with these investments, manual production has systematic limitations. First of all, intercoder reliability remains imperfect. For example, \textcite{nyrup2025paths} reports intercoder reliability of around 0.80 for cabinet-level biographical attributes, with agreement falling below 0.70 for certain variables.\footnote{Similar levels of coder disagreement are reported in other elite datasets, including LEAD \parencite{ellis2015introducing}, Archigos \parencite{goemans2009introducing}, and WhoGov \parencite{nyrup2020governs}, where resolving inconsistencies often requires multiple coding rounds or adjudication by senior researchers. More broadly, methodological surveys of text and manual data construction highlight that human hand-coding can introduce measurable errors and biases, motivating audits, cross-validation, and supervised approaches \parencite{grimmer2013text, gentzkow2019text}.} More importantly, most datasets remain static snapshots. Among the datasets surveyed, only a small fraction have been updated in the past five years, while widely used resources such as Archigos \parencite{goemans2009introducing} and LEAD \parencite{ellis2015introducing} have remained unchanged for a decade or more. Updating comprehensive elite datasets often requires thousands of additional labor hours, making continuous maintenance prohibitively costly. These production constraints shape the substantive scope of political inquiry. Existing datasets disproportionately focus on actors at the apex of political power, while mid-level bureaucrats, local officials, and other actors central to policy implementation remain largely absent from comparative data. Even among top-tier elites, coverage is uneven: while finance ministers \parencite{armstrong2024financial} and foreign ministers \parencite{back2021war} are relatively well documented, systematic data on portfolios such as education, health, or infrastructure remain scarce. 

The structure of available information further constrains what can be collected. When comprehensive Wikipedia biographies exist, researchers can extract structured facts from consolidated text. Such cases, however, are unevenly distributed across countries and levels of government and often omit early careers, family ties, or post-tenure activities. In their absence, researchers must reconstruct careers by manually searching across government websites, news archives, and organizational announcements. Information is then fragmented across heterogeneous sources, frequently in local languages and embedded in noise. As a result, empirical research tends to concentrate where information is easiest to obtain rather than where theoretical questions are most consequential, producing an ``information structure bias’’ \parencite{wilson2022geographical}.

Taken together, these realities expose two fundamental bottlenecks that any automated solution must address. The first is cost and scalability: manual coding scales linearly with dataset size, limiting expansion beyond narrow elite populations and hindering timely updates. The second is transparency and replicability: manually assembled datasets typically release only final records, with limited documentation of sources, conflicts, or adjudication rules, complicating verification and reuse. These constraints restrict not only the \textit{scale} of political data production but also its \textit{verifiability}, a growing concern as comparative political science increasingly relies on large-$N$ observational evidence.

The challenges discussed above extend beyond elite studies. Scholars studying contentious politics increasingly rely on real-time event data scraped from news sites and social media \parencite{king2013censorship,muthiah2015planned,zhang2019casm}. Legal and regulatory research requires tracking policy evolution across fragmented official gazettes, court databases, and agency announcements \parencite{baturo2017understanding,fang2025decoding,liebman2020mass}. Trade and investment research depends on synthesizing information from corporate filings, diplomatic cables, and industry publications \parencite{hassan2019firm,thrall2025informational}. In each case, the core challenge is identical: transforming vast, unstructured, and often conflicting information into valid, structured datasets at scales that manual coding cannot sustain. Political fact extraction, therefore, represents less a niche technical problem than a fundamental bottleneck constraining the empirical scope of comparative political science.

\section{From Classification to Extraction: The Context Challenge}

If manual coding is the bottleneck, recent advances in generative language models offer a theoretical solution. A rapidly growing body of work demonstrates that Large Language Models (LLMs) can replicate human judgments on classification tasks such as ideology scaling \parencite{wu2023large}, stance detection \parencite{benoit2025using,gilardi2023chatgpt}, and topic classification \parencite{ornstein2025train} with high reliability.\footnote{Empirical validations demonstrate that LLMs not only match but often exceed human performance on political text classification. \textcite{gilardi2023chatgpt} find that ChatGPT's zero-shot accuracy surpasses crowdworkers by approximately 25 percentage points on tasks involving stance, topics, and frame detection, while also achieving higher intercoder agreement than trained human coders. \textcite{benoit2025using} show that LLM ratings of party manifestos correlate with expert judgments at 0.87--0.92, reaching the upper bound of human expert agreement, with intra-LLM consistency typically exceeding 0.90 compared to human intercoder reliability of 0.3--0.5. Likewise, \textcite{wu2023large} demonstrate that LLM-generated ideology scores achieve test-retest correlations of 0.997 and better predict human perceptions of politician ideology than traditional behavioral measures. These studies establish that modern LLMs already deliver both superior accuracy and consistency on classification tasks.} These classification tasks share a common structure: they map bounded, pre-selected texts into finite label sets, holding the input document fixed. The model receives a well-defined text—a speech, a manifesto, a social media post—and must interpret its content according to a predefined codebook.

Political data extraction, however, represents a fundamentally different computational problem.\footnote{We provide a formal mathematical distinction between classification and extraction in Online Appendix~\ref{sec:appendix_formalization_classification_extraction}.} Unlike classification, which assigns labels to fixed texts, extraction entails actively searching for and reconstructing structured facts from vast, dispersed information environments where no single document contains complete information. This shift introduces four compounding challenges that classification benchmarks do not address.\footnote{In Natural Language Processing (NLP) literature, this class of problems is commonly described as \textit{open-domain slot filling} or \textit{complex information extraction} \parencite{angeli2015leveraging}. Unlike classification tasks with predefined label sets, these approaches aim to recover canonical attribute values from unstructured text under a specified codebook.} First, extraction is \textit{open-domain}: relevant entities, organizations, and position titles are not exhaustively enumerated ex ante, so the system must recognize and standardize an effectively unbounded set of possible answers rather than selecting from a fixed menu. Second, extraction exhibits high \textit{task complexity}: a single record (one official's career) requires answering many heterogeneous sub-questions (e.g., identity resolution, appointment dates, organizational affiliations, position titles, status flags), and reconciling contradictions across sources, rather than producing a single label. Third, extraction involves \textit{context dependency}: information retrieval is inherently path-dependent. Discovering one fact (e.g., an official served as ``Assistant Secretary at Commerce'') provides the contextual cue necessary to locate subsequent facts (e.g., searching for ``Commerce Assistant Secretary 2015'' rather than the initial broad query ``John Smith government'').\footnote{This dependency mirrors the challenge of \textit{multi-hop question answering} \parencite{yang2018hotpotqa}, where answering a query requires aggregating evidence from disjoint text segments. For biographical data, a resignation date in one document may resolve the tenure end date for a position mentioned in another, but only if both documents have been located and linked.} Unlike independent classification labels, career events form temporal sequences that require stateful reasoning to reconstruct.

These three challenges are formidable, and current evidence does not establish that LLMs can reliably address them at scale in real-world extraction scenarios. In principle, the same generative capabilities that enable classification could extend to these problems, yet translating this theoretical potential into validated extraction systems remains an open question. Even perfect solutions to the first three challenges would not resolve a fourth constraint that is orthogonal to model capability: the \textit{long-context problem}. In open-ended environments such as the open web, potentially relevant information for a single individual is dispersed across hundreds or thousands of documents (far exceeding the token budgets of even extended-context models), and is submerged in vast quantities of irrelevant content. This challenge operates on two dimensions, neither of which can be resolved through straightforward technical improvements.

First, models may fail to locate related information or generate false information to reconcile contradictory or misleading signals in too long contexts  \parencite{liu2024lost,mallen2023not,shi2023large}.\footnote{\textcite{liu2024lost} demonstrate a U-shaped performance curve in multi-document question answering: accuracy peaks when relevant information appears at the beginning or end of long contexts but degrades sharply (often by over 20 percentage points), when the same information is positioned in the middle. This ``lost-in-the-middle'' phenomenon persists across model families (GPT-3.5, Claude, open-source alternatives) and is not resolved by simply extending context windows: models with 16K-token capacity perform identically to their 4K counterparts when processing inputs that fit in both, indicating that raw capacity does not translate into robust information use \parencite[162]{liu2024lost}} Second, and more fundamentally, there exists a \textit{capacity constraint} that no architectural refinement can overcome: even models with 100K+ token windows cannot accommodate the full universe of potentially relevant documents for a given extraction target. A mid-level bureaucrat in a large country may be mentioned across thousands of government websites, news articles, and policy documents spanning decades; web searches routinely return result sets that, if naively concatenated, would exceed 1 million tokens. These findings reveal that the binding constraint in open-ended extraction is not merely reading long text, but deciding which sources to read and how to condense them into high-signal inputs before structured coding. Valid extraction at scale therefore requires an architectural solution that governs information selection and condensation \emph{before} LLM-based coding can be meaningfully applied.

\section{An Agentic Solution to the Extraction Challenge}

Building on the preceding analysis, we introduce an agentic architecture that directly targets the upstream bottleneck of evidence acquisition. The core idea is to decompose extraction into two analytically distinct stages: \textit{synthesis}, which locates, evaluates, and consolidates relevant evidence from open-ended sources, and \textit{coding}, which extracts structured facts from curated inputs. We operationalize this design through a recursive retrieval-and-synthesis loop that mirrors the iterative logic of human research and enables valid extraction from noisy web environments.

The central difficulty of open-web extraction lies in deciding \textit{which} sources to consult and \textit{how} to compress dispersed evidence into inputs suitable for downstream coding. Standard Retrieval-Augmented Generation (RAG) systems address this by retrieving document chunks based on semantic similarity to a query, then passing concatenated results to the model \parencite{lewis2020retrieval}. In a web search scenario, this is equivalent to issuing a single keyword query, retrieving the top-ranked pages, and conducting extraction on the aggregated results. This one-shot approach is fundamentally brittle under the \textit{context dependency} challenge identified earlier. The relevance of a document often cannot be determined ex ante but depends on information uncovered in prior retrieval steps. Key entities, affiliations, and career transitions are frequently discoverable only after intermediate facts have been established, rendering fixed retrieval strategies systematically incomplete.

\begin{figure}[htbp]
  \centering
  \includegraphics[width=\textwidth]{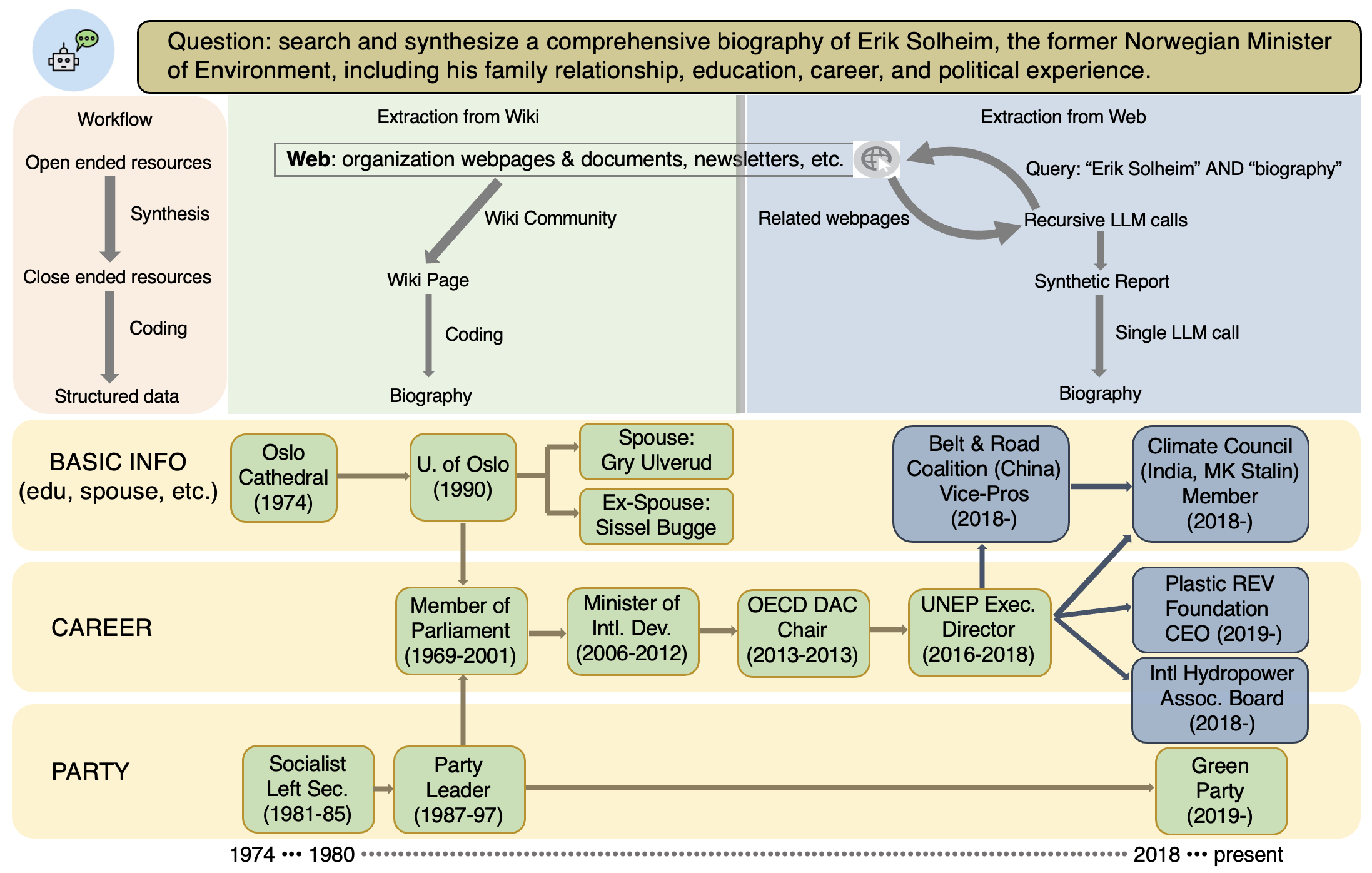}
  \caption{Two coding strategies for elite biographies. Left: when a Wikipedia page exists, we code directly from the curated page with a single LLM pass. Right: when Wikipedia is missing or incomplete, we search across web sources and iteratively synthesize a synthetic report, then code from that report. The lower panel illustrates the structured output as an ordered biography (career, education, and affiliations) anchored on a timeline. This contrast highlights why extraction from open-web sources requires adaptive synthesis rather than one-shot retrieval.}
  \label{fig:task_and_solution}
\end{figure}

Consider the biographical reconstruction task shown in Figure~\ref{fig:task_and_solution}. When Wikipedia contains a comprehensive biography (left panel, green boxes), a single LLM pass suffices to extract structured facts. But when Wikipedia is absent or incomplete (the common case for non-elite officials), the system must search across heterogeneous web sources (right panel, blue boxes). Crucially, informative follow-up queries are endogenous to what has been learned from earlier documents. Discovering that an official served as ``UNEP Executive Director'' provides the contextual anchor needed to locate subsequent positions (``Climate Council member''), professional affiliations (``Belt \& Road Coalition Vice-President''), or organizational roles (``Plastic REV Foundation CEO'') that would be invisible to an initial broad search. Static RAG, by committing to a fixed retrieval strategy before any evidence has been examined, cannot exploit these path-dependent cues.

RAG has emphasized the model’s capacity to \textit{observe}: given curated context, LLMs can reliably produce structured outputs in zero-shot or few-shot settings \parencite{ornstein2025train,gilardi2023chatgpt,benoit2025using}. What has received less attention is that modern LLMs can also \textit{act}. Specifically, they can generate executable commands that interact with external retrieval systems, enabling autonomous information gathering. In a recursive setting, the interleaving of action and observation allows the model to search iteratively, examine retrieved documents, reason about gaps in current knowledge, and decide what sources to consult next \parencite{yao2023react}. Each retrieval action is thus conditioned on information accumulated in previous steps, allowing the system to resolve the path-dependent nature of open-domain retrieval while progressively compressing a noisy information universe into a condensed corpus that avoids long-context constraints.

We operationalize this capability through an \textit{agentic framework} that repositions the LLM from passive reader to active research assistant. Rather than treating retrieval as preprocessing, the architecture implements a recursive reasoning–action loop: the model iteratively (i) \textit{reasons} about current knowledge gaps (e.g., ``I have identified the UNEP directorship but lack information on prior ministerial roles''), (ii) \textit{acts} to acquire missing evidence via targeted search queries or document inspection, and (iii) updates a running \textit{synthetic report} consolidating verified findings. Each step is executed through a minimal set of deterministic retrieval tools, which carry out machine-readable commands (e.g., \texttt{search(``Erik Solheim OECD DAC'')} or \texttt{$open_url(url\_5)$}) and return text for inspection. The agent iteratively incorporates evidence, decides whether further retrieval is needed, and maintains only the task description, recent interaction history, and current report in context to ground search decisions while avoiding context overflow. The final synthetic report, a compressed, wiki-like summary of curated evidence, serves as the sole input to the downstream coding step that produces the structured biography. The exact prompt templates used for the supervisor, searcher, and coder agents are listed in Online Appendix~\ref{sec:appendix_prompts}.

The agentic framework combines scalability, transparency, and validity in automated data production.\footnote{For a practical guide to applying this framework to new extraction tasks, including step-by-step recommendations on codebook design, synthesis configuration, and evaluation, see Online Appendix~\ref{sec:appendix_practical_guide}.} It can process thousands of targets in parallel without task-specific model training or human supervision, substantially reducing time and labor costs. At the same time, every retrieved source is archived and linked to the generated synthetic report, allowing researchers to inspect intermediate evidence, trace how claims were verified or adjudicated, and identify potential errors or biases \parencite{bail2024can}.\footnote{For a complete step-by-step trace of an agentic extraction run for a specific official, see the case study of Erik Solheim in Online Appendix~\ref{sec:appendix_case_study_full}.} To evaluate whether these architectural advantages translate into valid and scalable data production, we design a multi-stage empirical evaluation. Experiment~1 tests whether large language models can accurately code structured facts from curated biographical texts. Experiment~2 examines whether agentic synthesis from open-web sources can recover reliable biographical information in fragmented and noisy information environments. The last Experiment then assesses the architectural mechanisms underlying these results by holding the retrieved evidence fixed and varying how that evidence is represented to the coder. This diagnostic experiment clarifies why synthesis is essential for reliable extraction at scale.

\section{Experiment 1: The Coding Challenge}

This section addresses the first research question by isolating the \textit{coding} component of automated biography extraction. The objective is to assess whether, given identical and curated evidence, large language models can extract structured biographical facts with accuracy comparable to human coders. By holding the information environment constant and varying only the coder, this design directly tests whether coding itself constitutes a binding constraint in automated political data production. We implement this test in a setting where authoritative human-coded benchmarks exist, enabling direct validation of event-level extraction accuracy.

\subsection{Data}

\paragraph{Human-Coded Benchmark: The CPED}
The Chinese case provides a uniquely suitable benchmark for evaluating coding validity because it offers high-quality, human-coded biographical data at scale. Our analysis relies on the Chinese Political Elite Database (CPED), a comprehensive biographical database covering more than 4,000 key city-, provincial-, and national-level leaders since the late 1990s \parencite{jiang2018making}. Outside of the Central Organization Department's internal archives of the Chinese Communist Party, CPED is widely recognized as the most authoritative digital repository of Chinese political curricula vitae.\footnote{CPED provides detailed information on career trajectories, educational backgrounds, native place, birth year, ethnicity, and records of corruption investigations.} Importantly, all biographies in CPED are manually coded by trained research assistants following standardized rules, producing structured career histories that serve as a ground-truth benchmark for validation.

\paragraph{Sample Construction}
From the CPED population, we constructed a stratified random sample of 197 officials, balanced across three administrative ranks to capture variation in career complexity: (i) bureau-director level (equivalent to city mayors or provincial department heads), (ii) vice-ministerial level (provincial governors or vice-ministers), and (iii) ministerial level (provincial party secretaries or national ministers). This stratification ensures representation across the hierarchy of Chinese bureaucratic advancement, where career paths differ systematically by rank. These officials exhibit complex, longitudinal career histories typical of Chinese bureaucratic advancement, with multiple concurrent and sequential positions across party, government, and state-owned enterprise sectors. Decomposing these complex career histories into discrete positional observations yields over 4,000 structured biographical entries in the CPED benchmark. Each entry records a specific position with standardized fields: organization, location, role/title, start date, end date, and the administrative rank.

\subsection{Evaluation Design}

To isolate coding performance, human and LLM coders are provided with an identical information environment. For each official, the input consists of the full Baidu Baike profile associated with that individual.\footnote{In the Chinese context, official biographical information is highly standardized due to the party-state’s institutionalized \textit{nomenklatura} system. Baidu Baike, the dominant Chinese equivalent of Wikipedia, serves as the primary repository for official profiles. All sampled officials have Baidu Baike entries, which consolidate career narratives drawn from official announcements, government websites, and authoritative media sources. The human-coded ground truth in CPED was originally derived primarily from these same Baidu Baike profiles.} Under this setup, we generate two structured biographies per official. The baseline biography corresponds to the existing CPED record produced by trained research assistants. The treatment biography is generated by applying a long-context LLM coder to the identical Baidu Baike text in a single pass. Because both biographies draw on the same evidence source and follow the same codebook, any performance differences can be attributed to the coder rather than to variation in information availability or task definition. We evaluate multiple LLM architectures (Grok-4.1-Fast, Gemini-2.5-Flash, and Qwen-2.5) to assess robustness across models.

\paragraph{CGT Construction}

While CPED provides a high-quality human-coded benchmark, human annotation is not error-free. 
To establish a more reliable reference standard, we construct a \textit{Consolidated Ground Truth} (CGT) through a three-step validation pipeline. First, for each official, we pool all claims produced by both \texttt{Human\_wiki} and \texttt{LLM\_wiki} into a unified candidate set.\footnote{We use the subscript ``wiki'' as a generic shorthand to denote open-access, collaborative encyclopedia sources. For the Chinese sample, this refers specifically to Baidu Baike data; for the U.S. and OECD samples discussed later, it refers to Wikipedia.} These claims are then normalized into a standardized codebook \texttt{(entity, role, organization, start\_date, end\_date, status)}, enabling direct comparison across coders. Second, each normalized claim is subjected to evidence-based validation using an LLM-as-judge protocol \parencite{gu2024survey,li2024llms}, which evaluates supporting evidence from Baidu Baike and supplementary authoritative Chinese sources and classifies claims as \textit{verified}, \textit{contradicted}, or \textit{uncertain}. Verified claims enter the CGT; contradicted claims are excluded; uncertain cases are flagged for review. Third, to assess the reliability of automated validation, we conducted a manual audit of 500 randomly sampled claims (50 officials $\times$ 10 claims). Two independent Chinese-speaking research assistants reviewed the underlying evidence and judge classifications, achieving 94\% agreement. Identified systematic error patterns were corrected by refining validation prompts and re-running affected cases. Full CGT construction procedures, judge prompts, and audit results are documented in Online Appendix~\ref{sec:appendix_cgt}.

\paragraph{Performance Metrics}
We evaluate coding performance at the official level by comparing system-generated claims against the Consolidated Ground Truth (CGT). 
Let $\widehat{\mathcal{C}}_i$ denote claims produced by a candidate system for official $i$, and $\mathcal{C}^{\ast}_i$ denote CGT claims. We define true positives ($\text{TP}_i$), false positives ($\text{FP}_i$), and false negatives ($\text{FN}_i$) as:

\[
\text{TP}_i = \left|\widehat{\mathcal{C}}_i \cap \mathcal{C}^{\ast}_i\right|,\quad
\text{FP}_i = \left|\widehat{\mathcal{C}}_i \setminus \mathcal{C}^{\ast}_i\right|,\quad
\text{FN}_i = \left|\mathcal{C}^{\ast}_i \setminus \widehat{\mathcal{C}}_i\right|.
\]

From these quantities, we compute Precision, Recall, and F1 score:
\[
\text{Precision}_i = \frac{\text{TP}_i}{\text{TP}_i + \text{FP}_i},\quad
\text{Recall}_i = \frac{\text{TP}_i}{\text{TP}_i + \text{FN}_i},\quad
\text{F1}_i = \frac{2 \cdot \text{Precision}_i \cdot \text{Recall}_i}{\text{Precision}_i + \text{Recall}_i}.
\]

Precision captures the accuracy of extracted facts (the share of claims supported by verified evidence), while Recall measures coverage (the share of true career events successfully recovered). \textbf{F1} is the harmonic mean balancing both dimensions. High precision but low recall yields incomplete biographies; high recall but low precision contaminates datasets with hallucinations.

\paragraph{Estimation Strategy}

To estimate differences in coding performance between human and LLM coders, we fit additive fixed-effect models of the form:

\begin{equation}\label{eq:rq1_spec}
Y_i = \alpha + \beta \cdot \mathbf{1}(\text{Coder}_i = \text{LLM}) + \gamma \cdot \text{Model}_i + \epsilon_i,
\end{equation}

where $Y_i \in \{\text{F1}, \text{Precision}, \text{Recall}\}$ denotes the performance metric for official $i$. The indicator $\mathbf{1}(\text{Coder}_i = \text{LLM})$ captures whether the biography was produced by an LLM or by human coders, while $\text{Model}_i$ includes fixed effects for LLM architecture (Grok, Gemini, Qwen). The coefficient $\beta$ therefore identifies the average difference in coding performance between LLMs and humans, holding constant both the evidence source (Baidu Baike) and the extraction codebook (CPED). Standard errors are clustered at the official level, and 95\% confidence intervals are obtained via nonparametric bootstrap (1,000 iterations).

\subsection{Results}
\begin{figure}[htbp]
  \centering
  \includegraphics[width=0.9\textwidth]{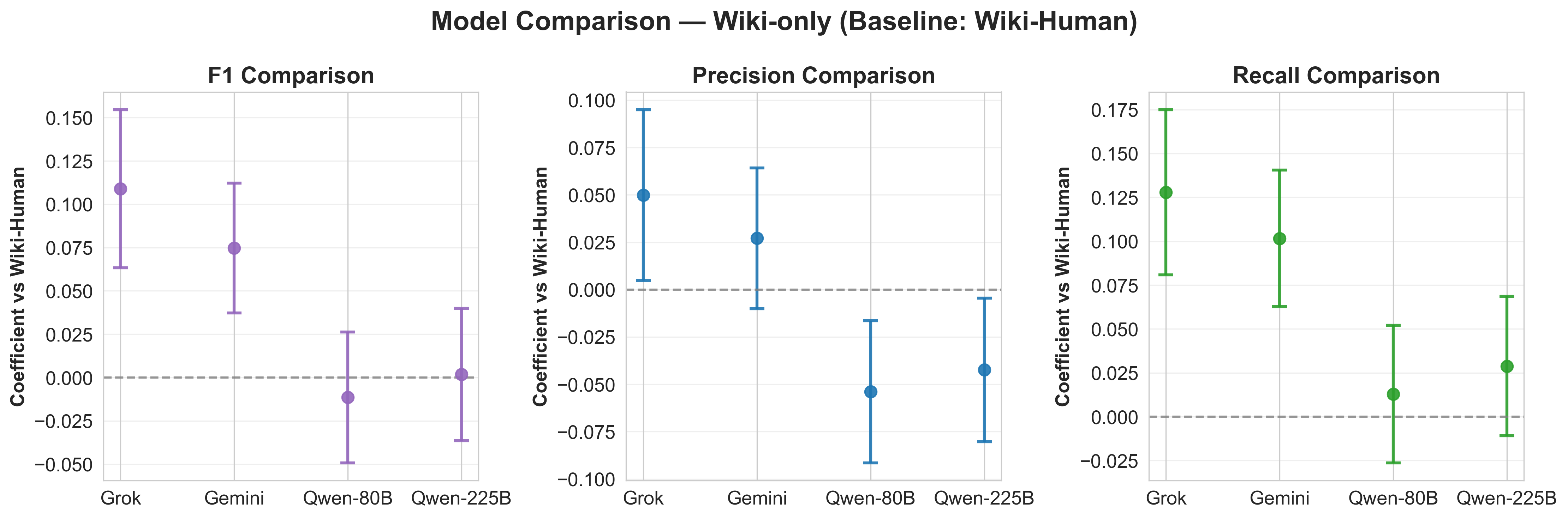}
  \caption{Experiment 1 Results: LLM coding performance relative to the human baseline (China sample, N=197). Points indicate coefficient estimates with 95\% confidence intervals. The human-coded baseline (\texttt{Human\_wiki}) is normalized to zero. Positive values indicate that LLMs outperform human coders on the corresponding metric.}
  \label{fig:rq1_results}
\end{figure}

Figure~\ref{fig:rq1_results} reports estimated differences in extraction performance relative to the human baseline, with coefficients from Equation~\ref{eq:rq1_spec} and 95\% confidence intervals. Across all metrics, contemporary LLM coders match or exceed the human baseline when applied to the same curated Baidu Baike corpus. Grok-4.1-Fast increases F1 by 0.109 significantly, driven by improvements in both precision and recall. Gemini-2.5-Flash exhibits similar, though more moderate gains with balanced improvements in precision and recall. Even smaller open-source models such as Qwen-3 achieve near-human capability.

These results establish that coding accuracy is not a binding constraint in automated political data production. When provided with curated inputs, LLMs can reliably map unstructured text to a complex, multi-field biographical codebook, achieving performance that matches or exceeds trained human coders. The most pronounced advantage lies in recall: leading models recover 10–16 percentage points more true career events than human coders working from the same corpus. This pattern is consistent with known limitations of manual annotation (principle agent problems, attention fatigue, selective reading, and time pressure), which lead human coders to systematically omit valid but less salient information, especially for officials with long and overlapping career histories. Importantly, these recall gains come with only modest changes in precision. 

Greater difference exists in marginal production costs. For human coding, we assume a skilled coder paid \$25 per hour. Given an average coding time of approximately 15 minutes per official, this yields a per-unit cost of \$6.25. For LLM coding, costs are computed based on token-level pricing for long-context inference. A typical Baidu Baike biography contains approximately 5,000–9,000 input and output tokens combined. Under the price of gemini-2.5-flash, the best-performing model, each official costs on average \$0.13 to process. 

\section{Experiment 2: The Synthesis Challenge}
Having established that modern LLMs can validly and efficiently code curated biographical inputs, we now turn to the central challenge addressed by this paper: whether automated systems can match or exceed human collective curation (Wikipedia) in the upstream task of consolidating noisy and fragmented web sources into codeable evidence. We evaluate this challenge using two complementary settings: contemporary U.S. political elites and ministerial officials from OECD countries. Unlike the Chinese case examined in Experiment~1, the U.S. and OECD contexts lack comprehensive human-coded biographical benchmarks covering the full range of elite career attributes.\footnote{Surprisingly, despite the size and maturity of the U.S. political science literature, there is, to the best of our knowledge, no publicly available dataset that provides CPED-style, career-long biographical coding for the full population of American political elites across offices and career stages. Existing resources typically focus on specific institutions (e.g., Congress or the presidency) or a narrow subset of attributes \parencite{bonica2016database}. For OECD countries, several cross-national databases document cabinet composition and tenure, but as shown in Table~\ref{tab:elite_datasets}, the scope of coded attributes remains substantially narrower than CPED, with limited coverage of education, pre-political careers, concurrent positions, and post-tenure trajectories.} Constructing such benchmarks manually would require thousands of research assistant hours and would reproduce precisely the scalability bottleneck that automated synthesis is designed to overcome. Rather than attempting to recreate human-coded benchmarks at prohibitive cost, Experiment~2 therefore evaluates synthesis performance in settings where only elite rosters are available ex ante and biographical information must be recovered from the open web. These contexts feature relatively high Wikipedia coverage, but with systematically varying degrees of information completeness and fragmentation beyond what encyclopedic curation captures. Together, they allow us to assess whether automated synthesis can recover biographical facts that are omitted, unevenly documented, or dispersed outside Wikipedia’s curated summaries.

\subsection{Data}
\paragraph{U.S. Political Elites}
The U.S. sample comprises 198 contemporary political elites drawn from a comprehensive roster compiled by the authors.\footnote{The full roster covers all state governors, members of Congress, Cabinet members, and Supreme Court justices from 1776 to 2025.} We employed stratified random sampling focused on the post-2000 period to ensure high data density on the open web, selecting three equal cohorts of 66 officials: Cabinet members, state governors, and members of the 119th Congress. Unlike the centralized personnel records found in authoritarian hierarchies, American political data is structurally decentralized. Information is dispersed across federal databases, state archives, and local media, requiring the synthesis agent to navigate a highly heterogeneous source landscape to reconstruct coherent biographical narratives.

\paragraph{OECD Political Elites}
The OECD sample consists of 200 ministerial officials from 36 member countries serving between 2011 and 2019, selected from the WhoGov 2.0 database \parencite{nyrup2020governs} via simple random sampling.\footnote{Unlike the China and U.S. samples, which use stratified sampling to capture vertical hierarchies, the OECD sample focuses on horizontally comparable cabinet-level officials to maximize cross-national coverage.} This sample tests the system's ability to handle breadth across linguistic and institutional contexts. The primary challenge here is not merely depth, but the unevenness of digital curation: officials from smaller member states or minor portfolios often lack English-language Wikipedia entries, forcing the agent to retrieve and synthesize information from native-language government websites, party manifestos, and local press.

\subsection{Evaluation Design}

Experiment~2 isolates the contribution of upstream synthesis by holding the downstream LLM coder fixed and varying only the method used to construct evidentiary corpora. All biography types are ultimately coded by the same LLM using an identical extraction codebook; differences in performance therefore reflect variation in how evidence is located, consolidated, and curated prior to coding.

\paragraph{Synthesis Conditions}

We compare three synthesis conditions that differ in how biographical evidence is assembled from the open web.

\textit{Human collective synthesis (Wikipedia baseline).}
Wikipedia represents the outcome of large-scale human collective curation: volunteer editors identify sources, resolve contradictions through citation norms, and compress verified information into narrative biographies. For this baseline condition, we use the existing Wikipedia page for each official \footnote{For wiki resources we used all sources with ``wiki'' domain; we also excluded grokipedia resources in all agent experiments.} as the sole input corpus. This setting reflects a best-case benchmark for human synthesis, benefiting from years of accumulated editorial effort. 

\textit{Agentic synthesis (full web).}
The full-web agent implements the iterative retrieval–reasoning loop described in Section~4. Starting from a broad query (official name and approximate role), the agent issues successive searches conditioned on information discovered in earlier steps, inspects retrieved documents, and consolidates verified claims into a running synthetic report. Wikipedia is treated as one source among many rather than an authoritative endpoint. Each claim in the report is explicitly linked to archived sources, and the agent terminates once sufficient evidence has been gathered to populate the extraction codebook. On average, the agent conducts 15–25 searches and 12–20 document inspections per official, with synthesis costs of approximately \$0.20 per case (search APIs), in addition to downstream coding costs. Full architectural details in Online Appendix~\ref{sec:appendix_architecture}.

\textit{Agentic synthesis without Wikipedia.}
To assess whether agentic gains depend on access to curated encyclopedic content, we implement a non-Wikipedia variant in which all wiki-domain URLs are blocked during retrieval and document inspection. The agent must reconstruct a Wikipedia-equivalent evidentiary base entirely from non-Wikipedia sources, such as government websites, parliamentary records, party materials, and news archives. This condition directly tests whether automated synthesis can mitigate information-structure bias in contexts where encyclopedic curation is sparse or absent.

\paragraph{Biography Types}

Combining these synthesis conditions with a fixed downstream LLM coder yields three biography types. The baseline biography (\texttt{LLM\_wiki}) is generated by applying the LLM coder to the Wikipedia page. The two treatment biographies are generated by applying the same coder to synthetic reports produced by the full-web agent (\texttt{LLM\_agent}) and the non-Wikipedia agent (\texttt{LLM\_nowiki}), respectively. Because the coder and extraction codebook are identical across conditions, performance differences isolate the contribution of upstream synthesis. Table~\ref{tab:rq2_design} summarizes the experimental contrasts and associated costs.

\begin{table}[htbp]
  \centering
  \caption{Experiment 2: Synthesis Method Comparison}
  \label{tab:rq2_design}
  \footnotesize
  \resizebox{\textwidth}{!}{%
  \begin{tabular}{@{}lccccc@{}}
  \toprule
  Biography Type & Synthesis Method & Downstream Coder & Corpus Type & Length & Cost \\
  \midrule
  \texttt{LLM\_wiki} (baseline) & Human (Wikipedia) & LLM (fixed) & Wiki page & $\sim$8k & \$0.01 \\
  \texttt{LLM\_agent} (treatment 1) & Agent (full-web) & LLM (fixed) & Synthetic report & $\sim$10k & \$0.21 \\
  \texttt{LLM\_nowiki} (treatment 2) & Agent (non-wiki) & LLM (fixed) & Synthetic report & $\sim$10k & \$0.21 \\
  \bottomrule
  \end{tabular}
  }
  \begin{minipage}{\textwidth}
  \footnotesize
  \vspace{0.4em}
  \textit{Notes.} All three biography types use the same downstream LLM coder (Grok-4.1-Fast). The experimental contrast isolates the synthesis contribution by holding the coder constant and varying only the upstream evidence construction method. Cost includes both synthesis (search API) and coding (LLM API) expenses per official.
  \end{minipage}
\end{table}

\paragraph{Ground Truth Construction}

As in Experiment~1, evaluation in Experiment~2 relies on a similar design by constructing \textit{Consolidated Ground Truth} (CGT) through evidence-based validation. For each official, we pool all claims extracted across synthesis conditions into a unified candidate set, normalize them into a common codebook, and validate each claim against archived source evidence using an automated judge. Verified claims constitute the CGT used for evaluation. To assess reliability, we conduct targeted manual audits on a random subset of claims across both the U.S. and OECD samples, including multilingual cases. Agreement between automated judgments and human review exceeds 90\%. Detailed consistency metrics for a randomly selected sample of officials across diverse linguistic contexts (OECD) are reported in Online Appendix Table~\ref{tab:audit_consistency}.

\paragraph{Performance Metrics}

Performance is evaluated using the same individual-level Precision, Recall, and F1 metrics as in Experiment~1, computed by comparing system-generated claim sets to the CGT.

\paragraph{Estimation Strategy}

We estimate the effect of agentic synthesis using the following specification:
\begin{equation}
\label{eq:rq2_spec}
Y_i = \alpha + \beta_1 \cdot \mathbf{1}(\text{Synthesis}_i = \text{Agent}) + \beta_2 \cdot \mathbf{1}(\text{Synthesis}_i = \text{NoWiki}) + \gamma \cdot \text{Controls}_i + \epsilon_i,
\end{equation}
where $Y_i \in \{\text{F1}, \text{Precision}, \text{Recall}\}$ denotes performance for official $i$. The baseline category is Wikipedia-based synthesis. Controls include downstream model indicators and sample fixed effects (U.S. vs.\ OECD). Coefficients $\beta_1$ and $\beta_2$ capture the average performance difference between agentic and human collective synthesis, holding the coder constant. Standard errors are clustered at the official level, with 95\% confidence intervals obtained via bootstrap.

\subsection{Results}
To rigorously quantify the synthesis challenge in these contexts, we first present the composition of URLs retrieved by our agentic framework across all three regions (including the China sample from Experiment 1 for comparison). Table~\ref{tab:url_composition_by_region} summarizes the distribution of these sources (see also Figure~\ref{fig:composition_comparison} in the Online Appendix for a visual breakdown). The data reveals distinct structural divergences across regions. While the average volume of retrieved URLs is consistent across samples ($\approx$ 21--22 URLs per official), the composition of evidentiary sources differs fundamentally. The China baseline exhibits a high concentration of state-sanctioned information, with 78.4\% of evidence derived from journalism (43.4\%) and official government sources (35.0\%). In contrast, the U.S. sample demonstrates a highly fragmented distribution. Reliance on official government sources drops to 26.9\%, while civil society sources, including non-wiki databases (10.4\%) and NGO groups (8.9\%), comprise a substantial portion of the evidence base, compared to negligible levels in China. The OECD sample occupies an intermediate position, balancing journalism (25.8\%) and government sources (22.1\%) with a significant reliance on wiki-based references (17.4\%). This contrast confirms that for democratic elites, valid extraction requires synthesizing evidence from a broad, diverse spectrum of non-official sources.

\begin{table}[htbp]
  \centering
  \caption{Retrieved URL Category Composition by Sample (Top 6 Categories)}
  \label{tab:url_composition_by_region}
  \footnotesize
  \begin{tabular}{@{}lccccccc@{}}
  \toprule
  Region & Avg URLs & Govt & Wiki & Journalism & Databases & NGO & Social \\
  \midrule
  China & 21.2 & 7.44 (35.0\%) & 2.07 (9.8\%) & 9.22 (43.4\%) & 0.05 (0.2\%) & 0.41 (1.9\%) & 0.13 (0.6\%) \\
  US & 22.4 & 6.02 (26.9\%) & 2.58 (11.5\%) & 3.10 (13.8\%) & 3.18 (14.2\%) & 1.99 (8.9\%) & 0.63 (2.8\%) \\
  OECD & 22.1 & 4.87 (22.1\%) & 3.02 (13.7\%) & 5.70 (25.9\%) & 1.63 (7.4\%) & 1.71 (7.8\%) & 1.15 (5.2\%) \\
  \bottomrule
  \end{tabular}
  \begin{minipage}{\textwidth}
  \footnotesize
  \vspace{0.4em}
  \textit{Notes.} Values represent the average number of unique URLs retrieved per official by the agentic framework, with the category's share of total regional volume in parentheses. Source categories are classified as follows: ``Govt'' includes official government sources; ``Wiki'' includes Wikipedia and wiki-derived encyclopedias; ``Journalism'' covers news media outlets; ``Databases'' refers to non-wiki structured reference databases (e.g., VoteSmart, Ballotpedia); ``NGO'' includes advocacy groups and NGOs; ``Social'' includes personal or professional social media platforms.
  \end{minipage}
\end{table}

\begin{figure}[htbp]
  \centering
  \includegraphics[width=0.9\textwidth]{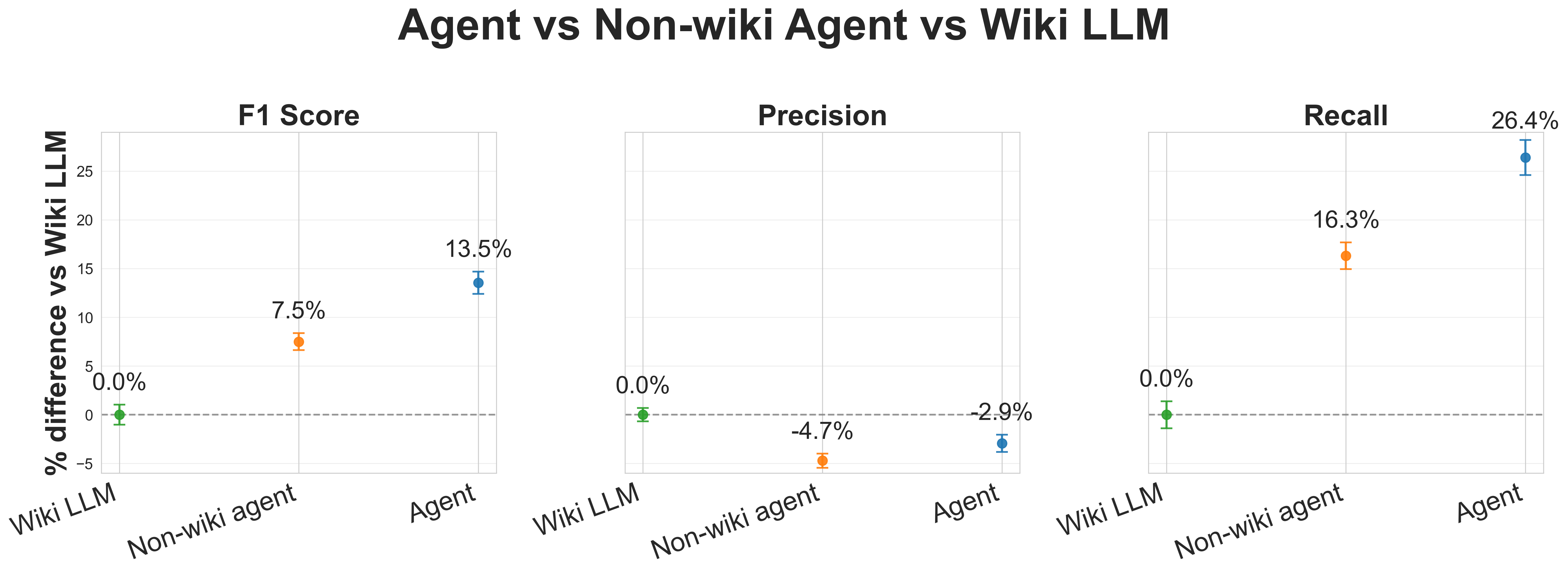}
  \caption{Agentic synthesis versus Wikipedia baseline (pooled U.S. and OECD samples, $N=398$). Points indicate coefficient estimates from Equation~\ref{eq:rq2_spec} with 95\% confidence intervals. The Wikipedia baseline (LLM\_wiki) is normalized to zero. Positive values indicate that agentic synthesis outperforms Wikipedia-based extraction.}
  \label{fig:rq2_aggregate}
\end{figure}

Figure~\ref{fig:rq2_aggregate} reports the aggregate effects of agentic synthesis relative to the Wikipedia baseline, pooling the U.S. and OECD samples ($N=398$).\footnote{For completeness, Appendix~\ref{sec:appendix_china_synthesis} reports parallel results for the China sample, where the encyclopedic baseline is already highly curated.} Across both contexts, agentic synthesis substantially improves coverage while maintaining acceptable levels of accuracy. The full-web agent (\texttt{LLM\_agent}) increases F1 by 14.7 percentage points (95\% CI: [12.1, 17.3]), driven primarily by large recall gains of 31.4 points (95\% CI: [27.8, 35.0]). Precision declines modestly by 5.2 points (95\% CI: [$-7.8$, $-2.6$]). Even when prohibited from accessing Wikipedia, the non-wiki agent (\texttt{LLM\_nowiki}) achieves sizable improvements, increasing F1 by 11.7 points (95\% CI: [9.3, 14.1]) and recall by 24.3 points (95\% CI: [21.1, 27.5]), with a comparable precision reduction of 4.1 points (95\% CI: [$-6.5$, $-1.7$]). Crucially, the dominance of recall gains underscores that agentic synthesis primarily serves to bridge information gaps left by human curation, significantly expanding the scope of biographical records beyond encyclopedic baselines.

\begin{figure}[htbp]
  \centering
  \includegraphics[width=0.9\textwidth]{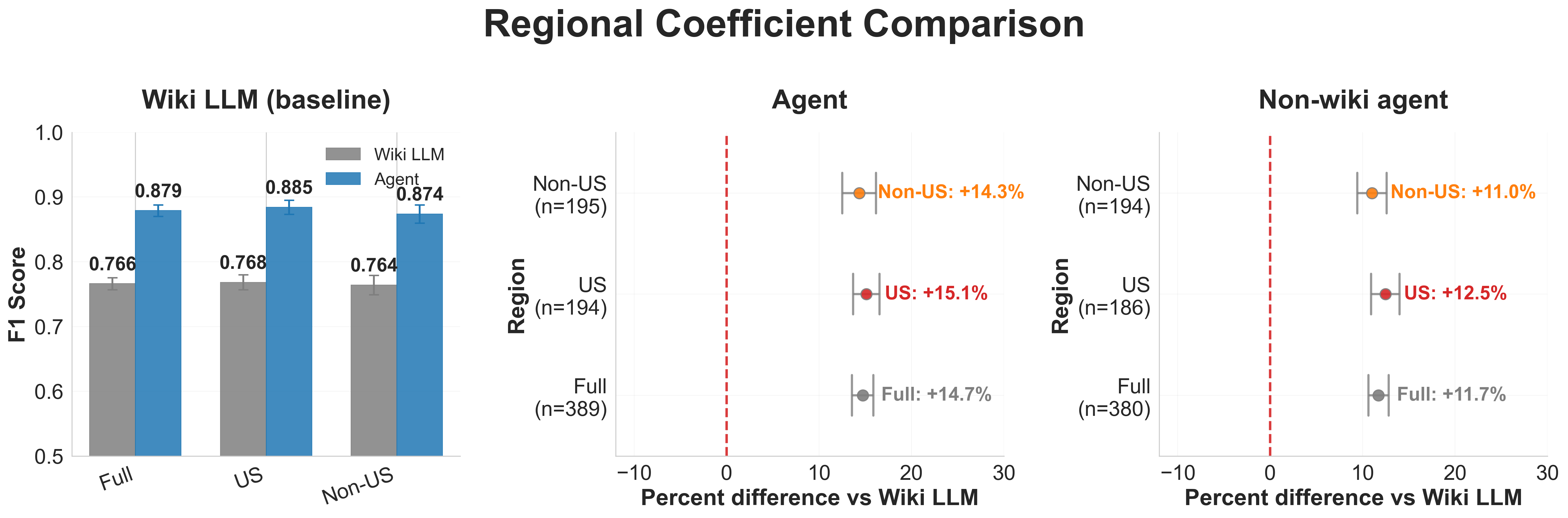}
  \caption{Agentic synthesis effects by sample. Points report F1 estimates with 95\% confidence intervals. Baselines are normalized to sample-specific Wikipedia means (U.S.: 0.82; OECD: 0.73).}
  \label{fig:rq2_by_sample}
\end{figure}

The regional decomposition in Figure~\ref{fig:rq2_by_sample} further clarifies the magnitude and generality of these gains. Across regions, agentic synthesis raises absolute F1 from approximately 0.76--0.77 under the Wikipedia baseline to roughly 0.87--0.89, a performance level that is well within the range required for downstream empirical applications. Importantly, these improvements are not driven by a single context. For the full-web agent, F1 increases by 15.1 percentage points in the U.S. sample and by 14.3 points in the non-U.S. sample, yielding a pooled gain of 14.7 points. Even when Wikipedia access is prohibited, the non-wiki agent delivers substantial improvements: F1 rises by 12.5 points in the U.S., 11.0 points in the non-U.S. sample, and 11.7 points overall. Two implications follow. First, the scale of improvement is remarkably stable across regions, despite large differences in baseline coverage and information structure. Second, the slightly larger gains outside the U.S. are consistent with the intuition that agentic synthesis is most valuable where curated coverage is weakest. Together, these results indicate that agentic workflows do not merely refine already well-documented cases, but systematically elevate extraction quality across heterogeneous information environments.

Collectively, these results demonstrate that agentic synthesis systematically expands the informational scope of elite biographies. The significant gains in recall reflect a structural characteristic of human curation, which prioritizes high-visibility and prominent roles, even though existing Wikipedia data remains highly precise. Consequently, granular details such as early-career positions, officials from smaller nations, non-political affiliations, and concurrent appointments are frequently compressed or omitted. By actively querying local news archives, government records, and organizational filings, agentic workflows successfully recover this ``long tail'' of politically relevant information.

The resulting trade-off between precision and recall is both modest and favorable. While absolute precision remains robust at 0.82–0.85, recall improves substantially by 24–31 percentage points. For most empirical research, capturing a significantly larger volume of verified facts justifies a marginal increase in noise, especially given the transparency of source archives and the potential for downstream validation. Across both samples, agentic synthesis achieves F1 scores between 0.87 and 0.94. These levels exceed reported human intercoder reliability in comparable elite datasets and are achieved at a fraction of the cost required for manual data collection.

\section{Experiment 3: Why Synthesis Matters?}
\label{sec:diagnostic_generalization}
The first two experiments establish two results: large language models can accurately code structured biographies when provided with curated evidence, and agentic retrieval can substantially expand coverage beyond Wikipedia. A natural follow-up question is whether explicit synthesis is still necessary once long-context models can ingest very large inputs. If a model can process hundreds of thousands of tokens, one might expect that simply concatenating all retrieved documents would suffice for accurate extraction. This section evaluates that assumption. Holding the underlying web evidence fixed, we show that extraction performance depends materially on how evidence is represented to the coder. In particular, long context alone does not eliminate omission and degradation errors. Instead, a synthesis step that compresses and organizes evidence into a signal-dense representation is critical for reliable extraction.

\subsection{Evaluation Design}
To isolate representational effects, we hold the downstream coding procedure fixed and compare alternative representations of the same retrieved web evidence. Using a fixed Grok-based agent retrieval trajectory, we construct two long-context corpora from identical underlying sources (Table~\label{tab:diagnostic_bio_types}) . The first is a raw internet corpus (\texttt{LLM\_raw}), defined as the direct concatenation of all retrieved documents in full. The second is a refined internet corpus (\texttt{LLM\_refined}), defined as a compressed, signal-dense representation composed of selected passages produced during the agentic reasoning loop. For comparison, we retain the Wikipedia-based long-context baseline (\texttt{LLM\_wiki}). This design holds the evidence universe constant and varies only the representation presented to the coder.

\begin{table}[htbp]
  \centering
  \caption{Representations used in the diagnostic comparison}
  \label{tab:diagnostic_bio_types}
  \footnotesize
  \resizebox{\textwidth}{!}{%
  \begin{tabular}{@{}lcccc@{}}
  \toprule
  Condition & Upstream evidence construction & Downstream coder & Representation & Typical context length \\
  \midrule
  \texttt{LLM\_wiki} & Human (Wikipedia) & LLM & Wiki narrative & $\sim$8k \\
  \texttt{LLM\_raw} & Agent (fixed trajectory) & LLM & Raw concatenation & $\sim$300k \\
  \texttt{LLM\_refined} & Agent (fixed trajectory) & LLM & Refined passages & $\sim$30k \\
  \bottomrule
  \end{tabular}
  }
  \begin{minipage}{\textwidth}
  \footnotesize
  \vspace{0.4em}
  \textit{Notes.} For \texttt{LLM\_raw} and \texttt{LLM\_refined}, the underlying retrieval trajectory is identical; only the representation supplied to the coder differs.
  \end{minipage}
\end{table}

We estimate biography-specific associations between recall and two mechanism proxies. The first captures a \textit{quantity channel}: context length, operationalized using token-length bins. The second captures a \textit{quality channel}: language composition, measured as the share of non-English tokens in the coding input. Formally, we estimate specifications of the form:
\[
\text{Recall}_i
= \alpha
+ \sum_{a \in \mathcal{A}} \mathds{1}(\text{Bio}_i = a)\cdot (\psi_a M_i)
+ \gamma \cdot \text{Controls}_i
+ \epsilon_i,
\]
where $M_i$ denotes the mechanism proxy and $\psi_a$ captures biography-specific slopes. This design allows us to diagnose whether long-context failures arise from scale effects, representation quality, or both.

\subsection{Results}
\begin{figure}[htbp]
  \centering
  \includegraphics[width=0.9\textwidth]{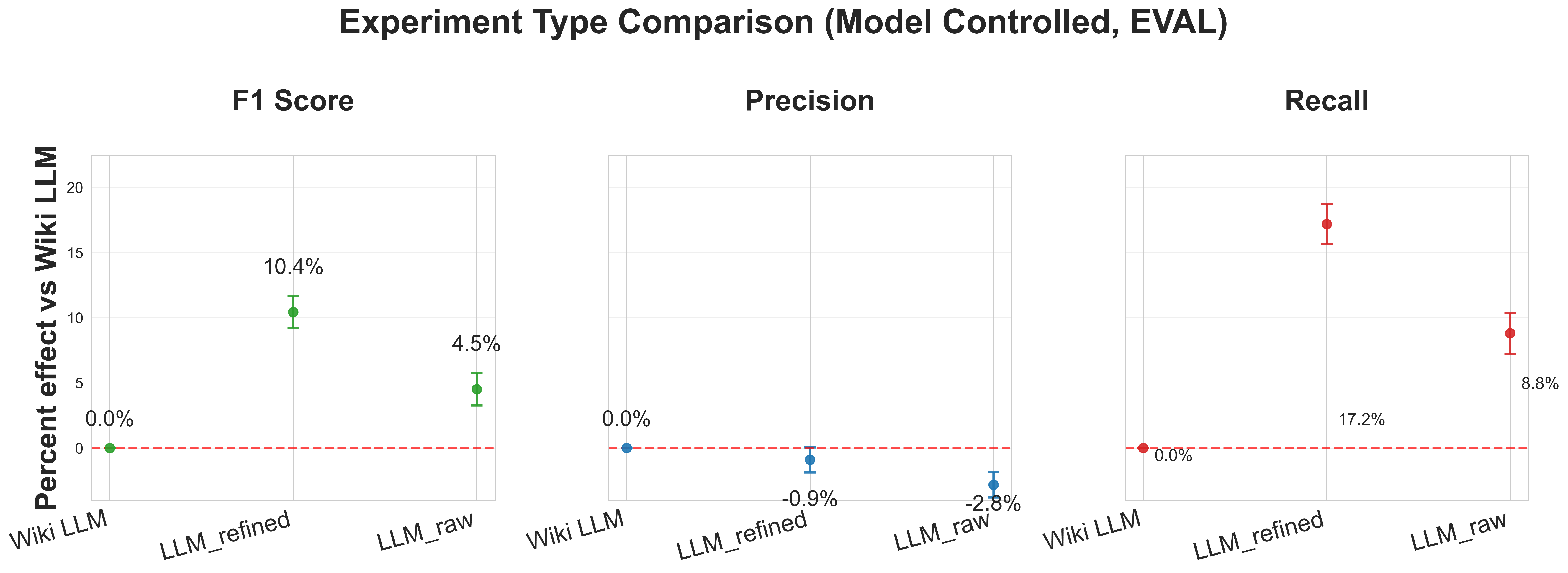}
  \caption{Refined versus raw corpora: the refinement premium.}
  \label{fig:refinement_premium}
\end{figure}

Figure~\ref{fig:refinement_premium} reports a clear refinement premium. Relative to the Wikipedia long-context baseline (\texttt{LLM\_wiki}), the refined representation (\texttt{LLM\_refined}) improves F1 by 10.4 percentage points and recall by 17.2 points, with only a modest precision change (-0.9 points). By contrast, raw concatenation (\texttt{LLM\_raw}) yields substantially smaller gains: F1 increases by 4.5 points and recall by 8.8 points, accompanied by a larger precision decline (-2.8 points). Because \texttt{LLM\_raw} and \texttt{LLM\_refined} are constructed from the same retrieved evidence, this contrast isolates representation as the binding factor. Long-context extraction failures therefore stem not from missing evidence, but from how evidence is organized and presented to the coder.

\begin{figure}[htbp]
  \centering
  \includegraphics[width=0.9\textwidth]{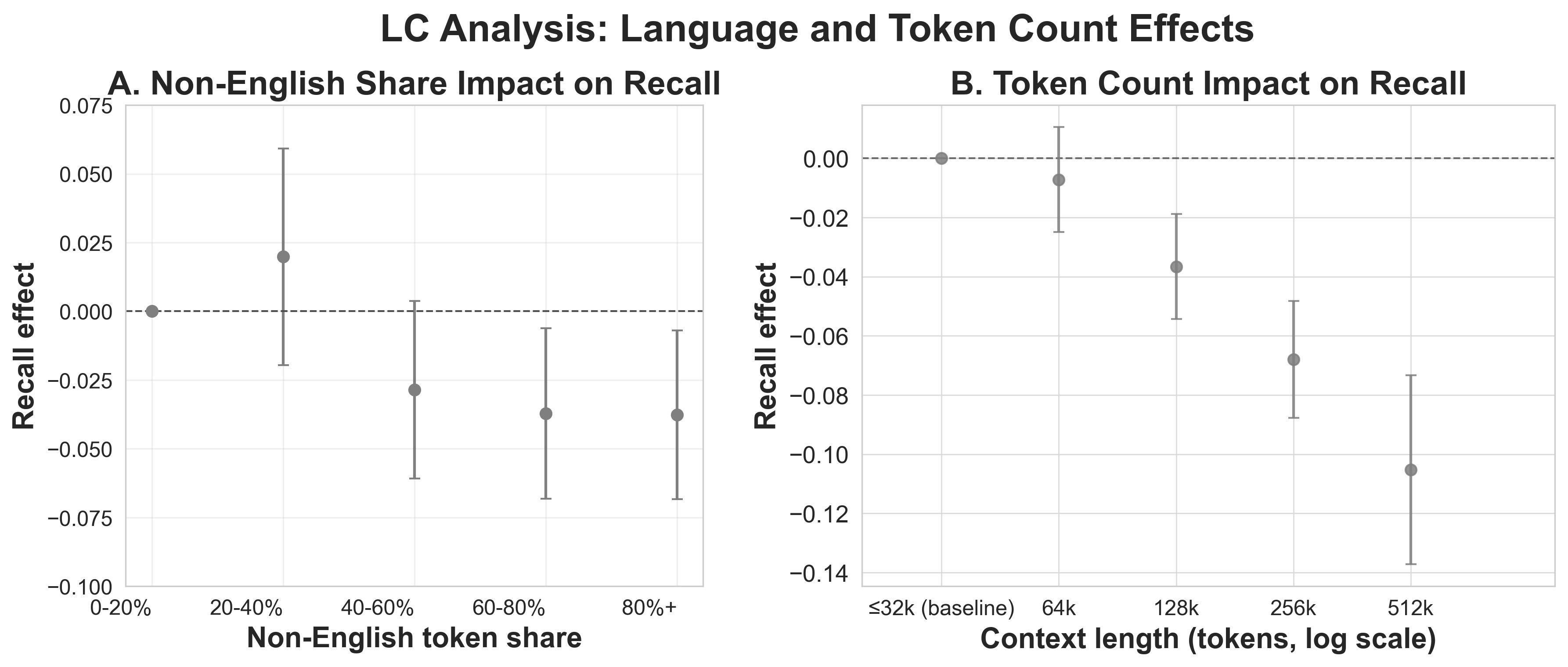}
  \caption{Mechanism evidence for long-context extraction failures. Panel A plots associations between non-English token share and recall; Panel B plots associations between context length and recall. Points denote estimated contrasts relative to the baseline bin; bars indicate 95\% confidence intervals.}
  \label{fig:lc_combined_regressions}
\end{figure}

Figure~\ref{fig:lc_combined_regressions} provides further mechanism evidence. Panel B shows a monotonic quantity penalty: recall declines as context length increases beyond moderate ranges, with the largest losses in the longest bins, consistent with long-context omission errors. Panel A shows a complementary quality channel: higher non-English token shares are associated with lower recall in several bins, indicating that heterogeneous or weakly structured inputs further strain extraction. Together, these patterns explain why synthesis-then-coding outperforms raw concatenation even when both draw on the same evidence universe.

These results clarify why explicit synthesis remains essential. Long context increases access to evidence, but it does not guarantee effective use of that evidence. Without refinement, large inputs dilute signal, exacerbate attention limits, and amplify representational noise. Agentic synthesis mitigates these failures by compressing evidence into structured, signal-dense representations that align with the extraction task. In short, synthesis is not a substitute for long context, nor is long context a substitute for synthesis. Reliable large-scale extraction requires both. Model heterogeneity under long-context conditions and descriptive comparisons of corpus composition, are reported in appendix \ref{sec:appendix_supplementary_results}.

\section{Conclusion}

``The historian,'' as \textcite{carr1961what} famously observed, ``is necessarily selective. The belief in a core of historical facts existing objectively and independently of the interpretation of the historian is a preposterous fallacy.'' A less noticed corollary applies with equal force to the political scientist: the structured datasets that undergird comparative inference are not neutral recordings of political reality, but artifacts of particular production processes---constrained by labor costs, source availability, and the cognitive limits of those doing the coding. This paper takes that constraint seriously and asks whether modern language technology can relax it without sacrificing validity.

Our answer, based on three experiments spanning Chinese, American, and OECD political elites, is cautiously affirmative---with important qualifications about how automated systems must be designed to earn that optimism. When given curated biographical inputs identical to those used by trained research assistants, contemporary large language models match or exceed human coding quality, with leading models recovering ten to sixteen percentage points more verified career events than their human counterparts working from the same source. In open-web environments, where no curated Wikipedia biography exists, an agentic synthesis workflow raises absolute F1 scores from the mid-seventies to the high eighties---performance levels that meet or exceed reported human intercoder reliability in comparable elite datasets---at roughly three percent of the per-unit cost of manual collection. And when we hold the evidence universe fixed and vary only how that evidence is represented to the coder, we show that long-context concatenation is not a substitute for synthesis: raw document aggregation yields substantially smaller and noisier gains than an explicitly refined, signal-dense representation derived from the same retrieved sources.

These results carry complicated normative implications for a field that has long treated human coding as the unquestioned benchmark for complex extraction. On one hand, the demonstration that LLMs can match and often exceed trained coders on identical inputs should prompt reflection about whether the costs of manual annotation have been buying the validity they were assumed to guarantee. Human coding is subject to well-documented failure modes---principal-agent attrition, selective reading, attention fatigue, and inconsistent adjudication across coders---that systematic LLM evaluation does not share to the same degree. On the other hand, automated pipelines introduce their own form of bias, one that is less visible precisely because it is technical rather than human. Our mechanism evidence shows that extraction performance degrades predictably under long inputs, multilingual corpora, and poorly structured source collections---biases that researchers may not notice if they do not inspect intermediate representations. The implication is that validity depends on upstream design choices (e.g., retrieval strategy, evidence compression, source credibility weighting) that are no less consequential than the coder's codebook.

The broader significance of our framework lies in what it makes newly possible rather than merely what it does more cheaply. As Table~\ref{tab:elite_datasets} documents, the most analytically important datasets in comparative elite research have remained static for years or decades, updated only when sustained institutional funding and coordinated RA labor can be mobilized. This structural rigidity shapes inquiry: researchers tend to study actors at the apex of formal institutions, in countries with dense English-language documentation, using attributes that require the fewest contextual inferences. The result is a systematic information structure bias in which the distribution of available data drives theoretical attention rather than the reverse \parencite{wilson2022geographical}. An agentic synthesis workflow does not eliminate this bias, but it substantially attenuates two of its main sources---the cost of expanding coverage to lower-visibility elites and the difficulty of maintaining data currency as political environments evolve. That the non-Wikipedia agent delivers F1 gains nearly as large as the full-web agent suggests that the technology is most valuable precisely where human curation is thinnest: officials from smaller states, minor portfolios, or less-documented institutional contexts where the conventional approach would simply leave the field blank.

There is also an interesting methodological parallel worth noting. The precision--recall trade-off we document in agentic synthesis mirrors a well-known dilemma in human coding between breadth and accuracy. Human coders working under time pressure tend to record the most salient positions and omit earlier or concurrent roles that require more inferential effort to recover. The agentic workflow reverses this asymmetry: it excels at recovering the ``long tail'' of biographical facts precisely because its search behavior is not anchored by salience but by evidential completeness. The modest precision declines we observe are consistent with false positives generated at the margin, but they are interpretable, source-traceable, and addressable through downstream validation in a way that human omission errors typically are not.

Several directions follow naturally from these findings. The relationship between extraction and classification in political science remains underexplored. Many constructs typically treated as classification targets---regime type, populism, democratic backsliding, policy diffusion---could in principle be reconceptualized as aggregations of extracted sub-claims: time-stamped events, actor attributions, institutional changes. Testing whether a synthesis-then-coding architecture improves both the accuracy and the auditability of such labels relative to direct prediction represents a promising frontier. A second direction concerns the contested end of the extraction spectrum. Our evaluations focus on relatively verifiable biographical facts where ground truth is unambiguous. Extending the framework to more interpretive attributes---rhetorical frames, policy positions, soft-power signals---raises harder questions about what a ``ground truth'' even means and how human and machine judgment should be combined when the target concept is itself contested. Finally, the biographical data produced by our framework create new empirical leverage for network approaches to elite politics. By standardizing career events, organizational affiliations, and temporal sequences across tens of thousands of officials, the resulting database permits reconstruction of fine-grained elite networks---co-service ties, overlapping tenures, shared educational institutions---at a scale and cross-national scope that hand-coded datasets have never been able to sustain. How these networks shape recruitment, policy coalitions, and regime stability remains largely unexplored in comparative work, not because the theoretical questions are unimportant, but because the data have not existed to answer them. They now can.

\singlespacing
\printbibliography

\clearpage
\end{refsection}

\clearpage
\setcounter{page}{1} 
\renewcommand{\thepage}{A-\arabic{page}}
\singlespacing

\appendix
\setcounter{section}{0}
\renewcommand{\thesection}{A\arabic{section}}

\section*{Online Appendix}

\setcounter{page}{0} 
\renewcommand{\thepage}{A-\arabic{page}}

\startcontents[sections] 
\renewcommand\contentsname{Table of Contents} 
\printcontents[sections]{l}{1}{\subsection*{\hspace*{1.2em}\contentsname}}{\setcounter{tocdepth}{2}}

\newpage
\begin{refsection}

\setcounter{page}{1} 
\renewcommand{\thepage}{A-\arabic{page}}

\setcounter{figure}{0}
\setcounter{table}{0}

\renewcommand{\thefigure}{\thesection.\arabic{figure}}
\renewcommand{\thetable}{\thesection.\arabic{table}}

\makeatletter
\@addtoreset{figure}{section}
\@addtoreset{table}{section}
\makeatother

\section{Formalizing Classification versus Extraction}
\label{sec:appendix_formalization_classification_extraction}

To formalize the distinction emphasized in the main text, we represent (i) classification as a closed-set mapping and (ii) political fact extraction as a two-stage pipeline that couples evidence synthesis with structured record construction. The key implication is methodological: larger or ``more capable'' language models need not yield reliable gains for extraction if the binding constraint is open-domain discovery, long-context integration, and multi-stage task execution under explicit budgets.

\subsection{Definitions and notation}
We use \textit{extraction} to denote the end-to-end task of producing structured political facts from unstructured corpora (a single document, a fixed document collection, or an open-ended document universe), following a predefined codebook. Within extraction, we distinguish \textit{synthesis} as evidence acquisition and refinement (search/browse/source selection, filtering, and condensation) from \textit{coding} as mapping a fixed, refined corpus into codebook-conformant structured records.

For classification, let $x$ denote the input text (a document or fixed set of documents) from an input space $\mathcal{X}$, and let $\mathcal{Y}$ denote a pre-defined, finite set of disjoint labels (e.g., $\mathcal{Y}=\{\text{Left},\text{Right}\}$). The classification task is a mapping
\[
f_{\text{classification}}:\mathcal{X}\rightarrow \mathcal{Y},
\]
often implemented by selecting the most likely label,
\[
\hat{y} = \arg\max_{y \in \mathcal{Y}} \Pr(y \mid x).
\]
Classification is therefore discriminative: the output space is fixed and known ex ante, and errors are primarily \emph{mislabeling}.

\subsection{Extraction as synthesis $\rightarrow$ coding}
Synthesis corresponds to evidence retrieval and refinement over a large, heterogeneous source universe (e.g., the open web). Let $\mathcal{D}$ denote the (implicit) universe of candidate documents and let $q$ denote a query (e.g., an entity name plus accumulated context from prior steps). Evidence retrieval can be written as a retrieval mapping that selects a bounded subset of evidence:
\[
f_{\text{retrieval}}:(q, h)\rightarrow \mathcal{D}_k \subseteq \mathcal{D},
\]
where $h$ is interaction history and $k$ is a number of retrieved corpus (or retrieved corpus). In practice, synthesis also includes filtering and condensation of $\mathcal{D}_k$ into a curated corpus that is feasible for downstream coding. Let $g$ denote a condensation operator that maps retrieved evidence into a curated corpus $x$ (e.g., a synthetic report) in an input space $\mathcal{X}$:
\[
g:\mathcal{D}_k \rightarrow \mathcal{X}.
\]
We therefore write synthesis as the composition
\[
f_{\text{syn}}(q,h) = g\!\left(f_{\text{ret}}(q,h)\right)\in \mathcal{X}.
\]

Coding then maps a fixed, refined corpus into structured records under an explicit codebook. Here, \text{syn} means synthesis. Let $\mathcal{B}$ denote a target codebook consisting of fields $\{b_1,\dots,b_K\}$ (e.g., Organization, Role, Start Date, End Date). Let $\mathcal{V}$ denote the vocabulary of the language model and let $\mathcal{V}^{\le L}$ denote the set of token sequences up to length $L$ (a convenient representation for LLM outputs). A codebook-conformant record is a tuple $z=(v_1,\dots,v_K)$ where each field value $v_k \in \mathcal{V}^{\le L}$. Let $\mathcal{Z}\subseteq (\mathcal{V}^{\le L})^K$ denote the set of such records, and let $\mathcal{T}$ denote the set of finite sequences of records (trajectories). We write the coding step as
\[
f_{\text{code}}:\mathcal{X}\times \mathcal{B}\rightarrow \mathcal{T}.
\]
The key difference from classification is that the output space is effectively open and the task is not separable: values are not drawn from a small closed set, and field-level decisions depend on other fields and on evidence scattered across documents. For elite biographies, the target is an ordered career trajectory rather than a single record. Let the extracted trajectory be $\hat{\tau}=(z_1,\dots,z_T)\in\mathcal{T}$, where each $z_t\in\mathcal{Z}$ is a codebook-conformant record. End-to-end extraction is the two-stage composition:
\[
\hat{\tau}=f_{\text{ext}}(q,h,\mathcal{B}) := f_{\text{code}}\!\left(f_{\text{syn}}(q,h),\mathcal{B}\right).
\]
Accordingly, beyond mislabeling, extraction failures include hallucination, span/value errors, missing events (recall loss), and codebook violations.

\subsection{Summary}
\begin{table}[htbp]
\centering
\caption{Conceptual distinction between classification and extraction tasks.}
\label{tab:appendix_cls_vs_ext}
\begin{tabular}{lll}
\toprule
\parbox[t]{0.15\textwidth}{Dimension} &
\parbox[t]{0.35\textwidth}{Classification ($f_{\text{cls}}$)} &
\parbox[t]{0.45\textwidth}{Extraction (Synthesis $\rightarrow$ Coding)} \\
\midrule
\parbox[t]{0.15\textwidth}{Pipeline} &
\parbox[t]{0.35\textwidth}{Single-step mapping} &
\parbox[t]{0.45\textwidth}{Two-stage: synthesis $f_{\text{syn}}$ (retrieval $f_{\text{ret}}$ + condensation $g$) then coding $f_{\text{code}}$; end-to-end $f_{\text{ext}}$} \\ \addlinespace
\parbox[t]{0.15\textwidth}{Output space} &
\parbox[t]{0.35\textwidth}{$\mathcal{Y}$ (finite, closed)} &
\parbox[t]{0.45\textwidth}{Intermediate evidence $\mathcal{D}_k \subseteq \mathcal{D}$, refined corpus $x\in\mathcal{X}$, and trajectory $\hat{\tau}\in\mathcal{T}$ (open, effectively unbounded)} \\ \addlinespace
\parbox[t]{0.15\textwidth}{Objective} &
\parbox[t]{0.35\textwidth}{Label selection} &
\parbox[t]{0.45\textwidth}{Evidence synthesis + reconstruction (recover structured values)} \\ \addlinespace
\parbox[t]{0.15\textwidth}{Typical errors} &
\parbox[t]{0.35\textwidth}{Mislabeling} &
\parbox[t]{0.45\textwidth}{Missed facts (recall loss); Unsupported facts (precision loss)} \\
\bottomrule
\end{tabular}
\end{table}

\section{Architecture and Model Details}
\label{sec:appendix_architecture}

This appendix documents the system architecture and model configuration used to produce the agentic biographies and the derived long-context corpora used in the experiments. In the main text, we emphasize the core conceptual feature---a tool-using, ReAct-style workflow for iterative evidence gathering and synthesis \parencite{yao2023react}. Here we provide additional implementation detail to make the design auditable and to clarify what is held constant across comparisons.

\begin{figure}[htbp]
  \centering
  \includegraphics[width=\textwidth]{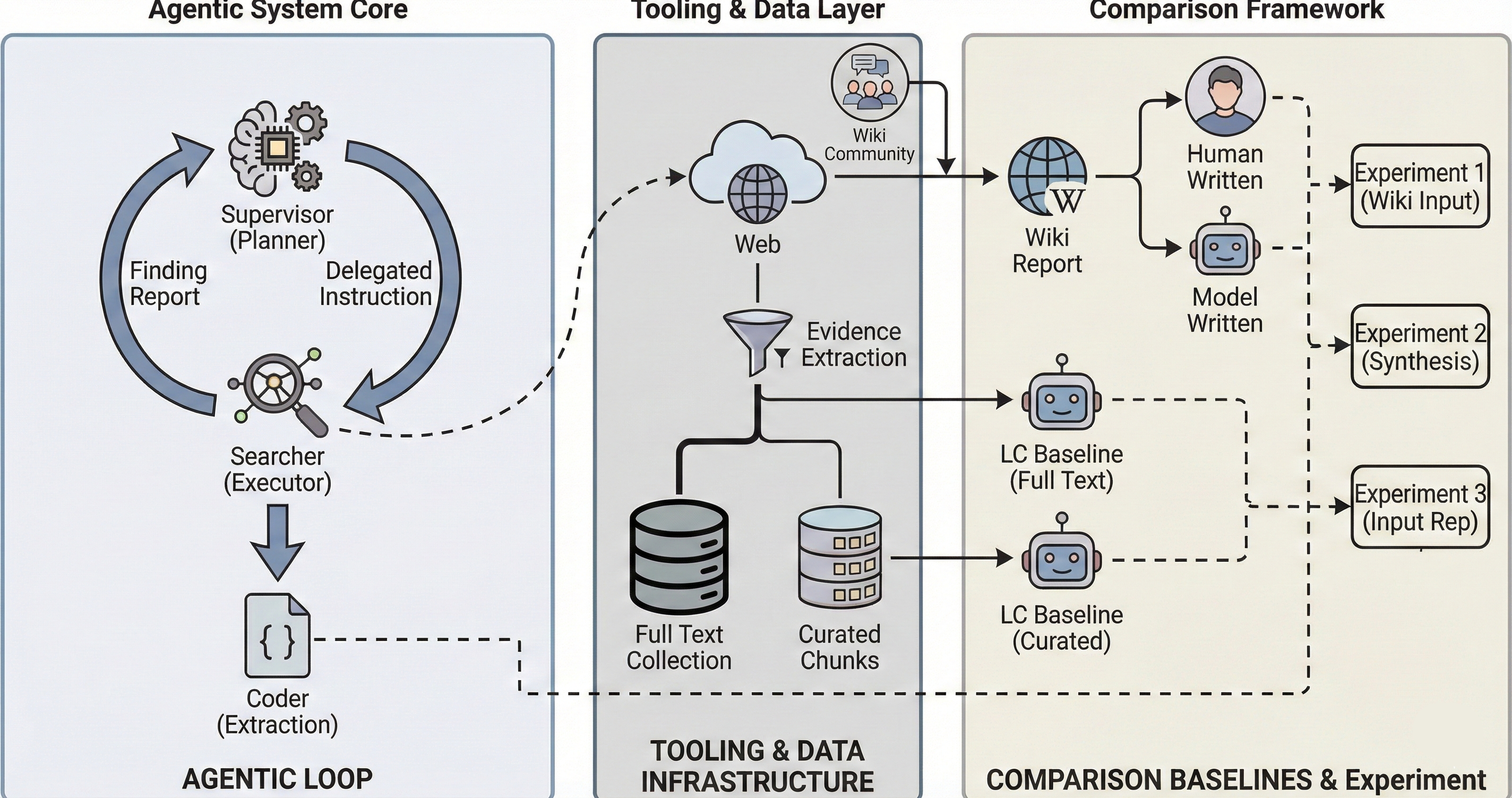}
  \caption{Architecture for agentic synthesis. A Supervisor maintains global state and delegates bounded retrieval tasks to specialized Searcher workers. Retrieved evidence is stored in an Archive and then mapped by a Coder into a structured biography.}
  \label{fig:appendix_les_code_architecture}
\end{figure}

\subsection{Architecture and information flow}
We implement a Supervisor--Worker architecture to manage the cognitive overhead of open-web synthesis. The central division of labor is between (i) strategic, long-horizon reasoning about what is missing and what to search next and (ii) tactical, short-horizon retrieval and reading of specific sources.

\paragraph{Core components.}
The \textbf{Supervisor} interprets the extraction objective and codebook, orchestrates the workflow over multiple cycles, and produces the final consolidated record. The Supervisor does not process the full raw web corpus directly; instead, it operates on structured evidence packets returned by specialized \textbf{Searcher} workers. The \textbf{Archive} stores retrieved content and provenance metadata (e.g., URLs and retrieval time), enabling deduplication, backtracking when contradictions arise, and transparent linkage between claims and supporting passages. The \textbf{Coder} converts the Supervisor’s stabilized draft and the archived evidence bundle into the structured biography output used for evaluation.

\paragraph{Two-stage pipeline.}
Our pipeline separates the extraction of political facts into (i) \textbf{synthesis}, which transforms a large, noisy document set into a higher-signal evidentiary representation, and (ii) \textbf{coding}, which maps that representation into a structured biography under a fixed codebook. This separation is crucial for interpretation: it allows us to hold the downstream coder constant while varying the upstream synthesizer (RQ2), and to isolate representation effects while holding the underlying web evidence fixed (Section~\ref{sec:diagnostic_generalization}).

\paragraph{Intermediate data structures.}
To keep multi-cycle synthesis auditable, we use explicit intermediate objects: (i) a \textbf{Structured Input} (objective, codebook, and constraints), (ii) a \textbf{System State} (running plan, partial biography, unresolved gaps, and bookkeeping), (iii) \textbf{Information Batch Overviews} produced by Searchers (source details, task-specific summaries, and extracted passages), and (iv) a \textbf{Structured Final Output} (codebook-conformant biography with evidence pointers via the Archive). Prompts are role-specific and enforce these interfaces; full prompt templates and tool specifications are provided in the replication materials.

\subsection{Models and operational constraints}
We evaluate multiple model families as downstream coders (e.g., \texttt{LLM\_wiki}) and as agentic components (e.g., \texttt{LLM\_agent}). The primary model families used in this version are \textbf{Grok-4.1-Fast}, \textbf{Gemini-2.5-Flash}, \textbf{Qwen-3-80B}, and \textbf{Qwen-3-225B}. Across experiments, we hold prompts and codebooks fixed within each role and keep system budgets (e.g., maximum steps and termination rules) constant within comparison Bios.

Model choice is consequential in agentic synthesis because end-to-end performance depends not only on reading comprehension, but also on tool-use reliability and the ability to sustain long-horizon interaction without drifting. We therefore prioritize models that jointly satisfy three practical constraints: strong reasoning in a fixed corpus, robust multi-turn planning and tool use, and affordability at scale. These constraints motivate the inclusion of ``Fast/Flash'' variants where available and efficient open-weight alternatives.

Finally, we design synthesis to remain operationally bounded. The Supervisor maintains a running search summary and a gap list, decomposes the task into Searcher instructions, and terminates when a step budget is reached or when the marginal value of additional retrieval declines. Because the workflow logs its actions and preserves archived evidence, we can audit intermediate representations and support claim verification during CGT construction (Appendix~\ref{sec:appendix_cgt}).

\begin{table}[htbp]
  \centering
  \caption{Agent Search Metrics by Model and Region}
  \label{tab:search_metrics}
  \footnotesize
  \setlength{\tabcolsep}{3pt}
  \begin{tabular}{@{}llrrrrr@{}}
  \toprule
  \textbf{Model} & \textbf{Version} & \textbf{Region} & \textbf{Officials} & \textbf{Searched (Avg)} & \textbf{Search Times (Avg)} & \textbf{Used URLs (Avg)} \\
  \midrule
  \multicolumn{7}{@{}l}{\textit{Gemini 2.5 Flash}} \\
  Gemini 2.5 Flash & non\_wiki & Overall & 398 & 62,389 (156.75) & 8,935 (22.44) & 9,299 (23.37) \\
  Gemini 2.5 Flash & non\_wiki & US & 198 & 30,672 (154.91) & 4,220 (21.31) & 4,917 (24.83) \\
  Gemini 2.5 Flash & non\_wiki & OECD & 200 & 31,717 (158.59) & 4,715 (23.57) & 4,382 (21.91) \\
  \midrule
  Gemini 2.5 Flash & model\_wiki & Overall & 398 & 61,046 (153.35) & 8,035 (20.18) & 7,075 (17.78) \\
  Gemini 2.5 Flash & model\_wiki & US & 198 & 28,917 (146.05) & 3,651 (18.44) & 3,348 (16.91) \\
  Gemini 2.5 Flash & model\_wiki & OECD & 200 & 32,129 (160.65) & 4,384 (21.92) & 3,727 (18.64) \\
  \midrule
  \multicolumn{7}{@{}l}{\textit{Grok-4.1-Fast}} \\
  Grok-4.1-Fast & non\_wiki & Overall & 398 & 68,677 (172.60) & 5,706 (14.34) & 10,020 (25.18) \\
  Grok-4.1-Fast & non\_wiki & US & 198 & 36,218 (182.92) & 2,992 (15.11) & 5,020 (25.35) \\
  Grok-4.1-Fast & non\_wiki & OECD & 200 & 32,459 (162.29) & 2,714 (13.57) & 5,000 (25.00) \\
  \midrule
  Grok-4.1-Fast & model\_wiki & Overall & 598 & 95,260 (159.28) & 5,938 (9.94) & 13,045 (21.82) \\
  Grok-4.1-Fast & model\_wiki & US & 198 & 30,230 (152.68) & 2,249 (11.36) & 4,433 (22.39) \\
  Grok-4.1-Fast & model\_wiki & China & 200 & 36,735 (183.68) & 1,522 (7.61) & 4,224 (21.12) \\
  Grok-4.1-Fast & model\_wiki & OECD & 200 & 28,295 (141.47) & 2,167 (10.84) & 4,388 (21.94) \\
  \midrule
  \multicolumn{7}{@{}l}{\textit{Qwen}} \\
  Qwen & model\_wiki & Overall & 398 & 41,411 (104.06) & 5,747 (14.43) & 8,640 (21.71) \\
  Qwen & model\_wiki & US & 198 & 21,225 (107.20) & 2,788 (14.08) & 4,646 (23.46) \\
  Qwen & model\_wiki & OECD & 200 & 20,186 (100.93) & 2,959 (14.79) & 3,994 (19.97) \\
  \midrule
  \multicolumn{7}{@{}l}{\textit{Qwen 225B}} \\
  Qwen 225B & model\_wiki & Overall & 398 & 35,302 (88.67) & 4,521 (11.36) & 7,121 (17.89) \\
  Qwen 225B & model\_wiki & US & 198 & 16,626 (83.97) & 2,204 (11.13) & 3,285 (16.59) \\
  Qwen 225B & model\_wiki & OECD & 200 & 18,676 (93.38) & 2,317 (11.59) & 3,836 (19.18) \\
  \bottomrule
  \end{tabular}
  \begin{minipage}{\textwidth}
  \footnotesize
  \vspace{0.4em}
  \textit{Notes.} This table presents detailed search metrics for the agent search process across different models, versions, and regions. Each metric column reports both the total count and the per-official average in parentheses. ``Searched (Avg)'' shows total search results retrieved (average per official). ``Search Times (Avg)'' shows the number of search operations performed (average per official). ``Used URLs (Avg)'' counts the total number of useful URLs retrieved by agent from search results (average per official). Overall region represents aggregated totals across all sub-regions (e.g., US + OECD, or US + China + OECD).
  \end{minipage}
\end{table}

Table~\ref{tab:search_metrics} documents the search behavior of different agent models during the upstream retrieval phase. For each model family (Gemini 2.5 Flash, Grok, Qwen, Qwen 225B), we report metrics across multiple configurations (non\_wiki and wiki variants) and geographic regions (Overall, US, OECD, China). The ``Searched'' column captures the total volume of search results processed, while ``Search Times'' indicates how many search queries each model issued. ``Used URLs'' counts the distinct web sources each model successfully retrieved, providing a proxy for retrieval breadth. These metrics reveal substantial variation in search strategies across models: some models (e.g., Gemini) issue more queries and retrieve more results, while others (e.g., Grok) converge more efficiently on relevant sources.

Notably, Grok demonstrates superior search efficiency across all models. As shown in Table~\ref{tab:search_metrics}, Grok achieves comparable or better coverage with substantially fewer search operations: it requires only 11.7 search times per official on average (across both versions), compared to 21.3 for Gemini 2.5 Flash and 14.4 for Qwen models. This efficiency translates directly into lower resource consumption and cost---Grok's per-official cost (\$0.15--\$0.18) is approximately 4--5$\times$ lower than Gemini 2.5 Flash (\$0.65--\$0.79) while maintaining competitive retrieval quality (88--102 unique URLs per official versus 123 for Gemini). Grok's ability to converge on relevant sources with fewer queries makes it particularly well-suited for large-scale agentic synthesis where cost and efficiency are critical constraints.

Larger models (Qwen 225B) tend to generate more output tokens than smaller variants (Qwen 80B), driving up costs despite similar input prices. The 225B variant v2 generates 5.2$\times$ more output tokens than the 80B variant (29.6M vs. 5.7M), contributing to its 2.9$\times$ higher total cost.

Table~\ref{tab:model_costs} reports the token usage and costs for each model configuration across different modes. Agent costs represent the full agentic synthesis pipeline (Searcher + Supervisor + Coder). Wiki LC (long-context) costs represent single-pass coding from Wikipedia pages. LC Raw and LC Synth are long-context variants using retrieved documents.

\subsubsection{Other infrastructure costs}
We relied on external providers for web research and robust web content retrieval. In particular, we used Jina and Exa as retrieval services capable of extracting full page contents from URLs. For web search, we used serp.dev to obtain programmatic access to Google’s search results.

Beyond model inference costs, production deployments should reserve budget for several API-related expenses. For search, about \$400 covers roughly 400,000 queries (around \$0.001 per request). Large-scale document fetching via Jina and exa services is on the order of \$200. Overall, our full experimental effort—including development, testing, experimentation, and evaluation—amounted to approximately \$5,000 in total spending across all LLM APIs, search, and retrieval services, as well as coders' hiring cost.

\begin{table}[htbp]
  \centering
  \caption{Model Token Usage and Costs}
  \label{tab:model_costs}
  \footnotesize
  \setlength{\tabcolsep}{4pt}
  \begin{tabular}{@{}llrrrrrr@{}}
  \toprule
  \textbf{Model} & \textbf{Mode} & \textbf{Input (M)} & \textbf{Output (M)} & \textbf{Input (\$)} & \textbf{Output (\$)} & \textbf{Total (\$)} & \textbf{Per Official (\$)} \\
  \midrule
  \multicolumn{8}{@{}l}{\textit{Agent Models}} \\
  Grok 4 Fast & Agent Wiki & 271.7 & 13.1 & \$54.35 & \$6.54 & \$60.88 & \$0.153 \\
  Grok 4 Fast & Non-Wiki & 323.4 & 15.0 & \$64.68 & \$7.49 & \$72.17 & \$0.181 \\
  \midrule
  Gemini 2.5 Flash & Agent Wiki & 754.5 & 12.3 & \$226.36 & \$30.79 & \$257.15 & \$0.646 \\
  Gemini 2.5 Flash & Non-Wiki & 938.2 & 12.7 & \$281.46 & \$31.69 & \$313.15 & \$0.787 \\
  \midrule
  Qwen3 225B & Agent Wiki & 517.4 & 29.6 & \$93.12 & \$15.99 & \$109.11 & \$0.274 \\
  \midrule
  Qwen3 80B & Agent Wiki & 341.4 & 5.7 & \$30.72 & \$6.25 & \$36.97 & \$0.093 \\
  \midrule
  \multicolumn{8}{@{}l}{\textit{Long-Context (LC) Modes}} \\
  Grok 4 Fast & Wiki & 9.0 & 1.3 & \$1.79 & \$0.64 & \$2.43 & \$0.006 \\
  Grok 4 Fast & LC Raw & 51.2 & 1.4 & \$10.23 & \$0.69 & \$10.92 & \$0.027 \\
  Grok 4 Fast & LC Synth & 23.4 & 1.8 & \$4.68 & \$0.88 & \$5.56 & \$0.014 \\
  \midrule
  Gemini 2.5 Flash & Wiki & 13.0 & 0.6 & \$3.90 & \$1.59 & \$5.49 & \$0.014 \\
  Gemini 2.5 Flash & LC Raw & 45.3 & 0.8 & \$13.59 & \$1.98 & \$15.57 & \$0.039 \\
  Gemini 2.5 Flash & LC Synth & 25.3 & 0.8 & \$7.60 & \$2.04 & \$9.64 & \$0.024 \\
  \midrule
  Qwen3 225B & Wiki & 13.0 & 2.5 & \$2.34 & \$1.37 & \$3.71 & \$0.009 \\
  Qwen3 225B & LC Raw & 29.5 & 2.0 & \$5.31 & \$1.06 & \$6.37 & \$0.016 \\
  Qwen3 225B & LC Synth & 24.3 & 2.4 & \$4.37 & \$1.28 & \$5.65 & \$0.014 \\
  \midrule
  Qwen3 80B & Wiki & 12.9 & 0.5 & \$1.16 & \$0.51 & \$1.67 & \$0.004 \\
  Qwen3 80B & LC Raw & 29.5 & 3.3 & \$2.65 & \$3.61 & \$6.26 & \$0.016 \\
  Qwen3 80B & LC Synth & 24.4 & 4.1 & \$2.19 & \$4.48 & \$6.67 & \$0.017 \\
  \bottomrule
  \end{tabular}
  \begin{minipage}{\textwidth}
  \footnotesize
  \vspace{0.4em}
  \textit{Notes.} Sample size: N=398 officials for Agent modes; N varies for LC modes. Agent Wiki includes models with Wikipedia access; Non-Wiki blocks Wikipedia during synthesis. Wiki represents single-pass long-context coding from Wikipedia pages only. LC Raw uses retrieved documents; LC Synth uses supervisor-enhanced retrieved documents. Input/Output in millions (M) of tokens. Price based on openrouter model prices.
  \end{minipage}
\end{table}

Table~\ref{tab:url_types} breaks down the average number of URLs retrieved per official by type, model, and region. We categorize URLs into eight types using a politician-centric reliability hierarchy: official government sources (primary/authoritative), wiki pages, news/journalism media (tertiary/interpretive), non-wiki reference databases (secondary/structured), social media platforms, NGO/advocacy sources, commercial sources, and other sources. The distribution varies substantially across models and regions. Notably, wiki variants consistently retrieve more wiki pages (2.16--4.17 per official) compared to non-wiki variants (1.36--2.04), while non-wiki variants rely more heavily on non-wiki reference databases (e.g., Grok non-wiki: 2.74 per official in US vs Grok v7 wiki: 2.34). Grok v7 (wiki) shows particularly high government source usage in China (7.44 per official, representing 35\% of all URLs) compared to US (26.9\%) and OECD (22.1\%). China data reflects only the final Grok v7 experiment due to processing log loss for earlier China search results.

\begin{table}[htbp]
  \centering
  \caption{URL Types per Official by Model and Region}
  \label{tab:url_types}
  \footnotesize
  \setlength{\tabcolsep}{3pt}
  \resizebox{\textwidth}{!}{%
  \begin{tabular}{@{}llrrrrrrrrr@{}}
  \toprule
  \textbf{Model} & \textbf{Region} & \textbf{Total} & \textbf{Govt} & \textbf{News} & \textbf{Wiki} & \textbf{Reference} & \textbf{Platforms} & \textbf{NGO} & \textbf{Commercial} & \textbf{Other} \\
  \midrule
  \multicolumn{11}{@{}l}{\textit{Grok}} \\
  Grok non-wiki & US & 25.61 & 9.02 & 3.80 & 0.18 & 4.04 & 0.71 & 2.28 & 0.78 & 4.81 \\
  Grok non-wiki & OECD & 25.00 & 7.56 & 7.27 & 0.34 & 1.94 & 1.25 & 2.12 & 1.28 & 3.24 \\
  \midrule
  Grok wiki & US & 22.39 & 6.02 & 3.10 & 2.58 & 3.18 & 0.63 & 1.99 & 0.53 & 4.35 \\
  Grok wiki & China & 21.23 & 7.44 & 9.22 & 2.07 & 0.05 & 0.13 & 0.41 & 0.00 & 1.92 \\
  Grok wiki & OECD & 22.05 & 4.87 & 5.70 & 3.02 & 1.63 & 1.15 & 1.71 & 1.23 & 2.73 \\
  \midrule
  \multicolumn{11}{@{}l}{\textit{Gemini}} \\
  Gemini non-wiki & US & 25.09 & 7.04 & 3.37 & 0.02 & 4.95 & 1.21 & 2.77 & 1.02 & 4.72 \\
  Gemini non-wiki & OECD & 22.02 & 4.62 & 5.43 & 0.03 & 2.78 & 2.20 & 2.11 & 1.86 & 2.99 \\
  \midrule
  Gemini wiki & US & 16.99 & 3.84 & 1.79 & 2.58 & 2.86 & 0.66 & 1.40 & 0.74 & 3.12 \\
  Gemini wiki & OECD & 18.92 & 3.43 & 3.79 & 3.03 & 1.91 & 1.67 & 1.50 & 1.48 & 2.11 \\
  \midrule
  \multicolumn{11}{@{}l}{\textit{Qwen 225B}} \\
  wiki & US & 16.85 & 4.59 & 2.24 & 2.65 & 2.35 & 0.83 & 1.58 & 0.19 & 2.41 \\
  wiki & OECD & 19.18 & 5.18 & 4.76 & 2.48 & 1.37 & 1.69 & 1.25 & 0.46 & 1.99 \\
  \midrule
  \multicolumn{11}{@{}l}{\textit{Qwen 80B}} \\
  wiki & US & 23.46 & 6.71 & 3.85 & 2.68 & 2.74 & 1.51 & 1.76 & 0.49 & 3.73 \\
  wiki & OECD & 19.97 & 4.59 & 2.81 & 2.77 & 2.06 & 2.98 & 1.39 & 0.76 & 2.60 \\
  \bottomrule
  \end{tabular}%
  }
  \begin{minipage}{\textwidth}
  \footnotesize
  \vspace{0.4em}
  \textit{Notes.} This table shows the average number of URLs retrieved per official by type, model, and region. Grok v7 = model\_wiki variant; Grok non-wiki = non\_wiki variant; Gemini v2 = model\_wiki variant; Gemini non-wiki = non\_wiki variant; Qwen 225B v2 and Qwen 80B = model\_wiki variants. China data reflects only the final Grok v7 experiment due to processing log loss for earlier China search results. Govt = official government sources; News = journalism and media; Wiki = Wikipedia and wiki-style pages; Reference = non-wiki reference databases (e.g., VoteSmart, Ballotpedia); Platforms = social media platforms; NGO = advocacy/NGO sources; Commercial = commercial/business sources; Other = uncategorized sources including entertainment, media, search engines, and miscellaneous.
  \end{minipage}
\end{table}


\section{Consolidated Ground Truth (CGT) Construction}
\label{sec:appendix_cgt}

This appendix provides the full claim-level CGT protocol summarized in the main text. Our goal is to produce a defensible, auditable reference set of claims for scoring.

\subsection{Protocol (pooling, consensus, and verification)}
\paragraph{Inputs (fixed pool per individual).}
For each individual $i$, we construct a fixed pool of \textbf{9 biographies}: four agent biographies (two agent model families $\times$ two variants), four \texttt{LLM\_wiki} biographies (four coder models applied to the same Wiki corpus), and one human-written Wiki biography (\texttt{Human\_wiki}). We construct the CGT from this pool and score all candidate systems against it. Long-context baselines (\texttt{LLM\_raw} and \texttt{LLM\_refined}) are scored against the CGT but are not included in the CGT pool to avoid mechanically altering the consensus set.

\paragraph{Step 1 (claim extraction and normalization).}
We decompose each biography into a set of atomic, comparable claims (e.g., education events; offices held with dates; party membership). We then normalize claims to reduce superficial disagreement: we canonicalize entity names and common aliases when available; standardize role and organization strings (e.g., ministry/agency names); and harmonize date formats, resolving partial dates into comparable intervals when possible. After normalization, paraphrases that express the same event are treated as the same claim.

\paragraph{Step 2 (consensus filter for high-confidence claims).}
For each normalized claim, we compute its presence rate in the 9-biography pool:
\[
presence(\text{claim}) = \frac{\#\{\text{bios containing claim}\}}{9}.
\]
Claims with $presence \ge 5/9$ enter the CGT as \textbf{high-confidence} claims. Claims with $presence \le 4/9$ are treated as \textbf{disputed/low-confidence} and proceed to evidence verification. 

\paragraph{Step 3 (evidence-conditional verification for low-confidence claims).}
Low-confidence claims are evaluated against a pooled evidence bundle: the union of archived passages and sources collected across all agent runs and variants for the same individual. We add a low-confidence claim to the CGT only if it is supported by explicit evidence in this pooled archive. Pooling across agent runs mitigates dependence on any single retrieval trajectory and improves robustness to idiosyncratic search failures. In our implementation, we operationalize this step with an evidence-conditional verifier (GPT-5-mini), which receives the candidate claim and the pooled archive text and returns a supported/unsupported judgment. In practice, we used the soft label to let LLMs label the level of support by 1-5, and treat claims scored above 3 as supported claims.

\paragraph{Step 4 (CGT definition and scoring).}
Let $\mathcal{C}^{\text{High}}_i$ denote the set of high-confidence claims and $\mathcal{C}^{\text{Validated}}_i$ the set of evidence-validated low-confidence claims. We define the claim-level CGT as:
\[
\mathcal{C}^{\ast}_i = \mathcal{C}^{\text{High}}_i \cup \mathcal{C}^{\text{Validated}}_i.
\]
Each candidate system output is converted into a normalized claim set $\widehat{\mathcal{C}}_i$ and scored against $\mathcal{C}^{\ast}_i$ using precision, recall, and F1 as defined in the main text. 

\subsection{Audit checks}
\label{sec:audit_checks}

We validate the reliability of the automated CGT construction through two distinct audit studies. To assess whether evidence-conditional verification aligns with expert judgment and external search verification, we drew a random sample of 20 officials from the OECD dataset. Table~\ref{tab:audit_consistency} summarizes the consistency rates, measured as the percentage of exact matches between the automated CGT verdicts and the alternative verification methods.

\paragraph{Human--machine alignment.}
Two graduate research assistants independently verified extracted claims for the sampled officials. Auditors were provided with the claim text and access to open-web search but were blinded to the model's verdict. For non-English sources, auditors utilized translation tools alongside original source inspection. Agreement rates were calculated by comparing human judgments against the automated judge's outputs. As shown in Table~\ref{tab:audit_consistency}, the average agreement rate is 91.3\%, indicating strong alignment between the automated verifier and human judgment in adjudicating low-consensus discoveries.

\paragraph{External validation (Exa).}
To rule out model-specific artifacts in the retrieval process, we cross-validated claims using Exa deepsearch, a neural search engine optimized for semantic retrieval and fact checking \parencite{exadeepsearch}. We queried Exa to retrieve high-quality, independent evidence for each claim and compared its verification results with our pipeline's verdicts. The analysis reveals a negligible discrepancy rate (average agreement 98.7\%), confirming that the synthesized ground truth is factually grounded and robust to retrieval method variations.

\begin{table}[htbp]
\centering
\footnotesize
\caption{Audit results: Consistency checks across verification methods for sampled OECD officials.}
\label{tab:audit_consistency}
\renewcommand{\arraystretch}{1.1}
\setlength{\tabcolsep}{4pt}
\begin{tabular}{@{}lllcc@{}}
\toprule
& & & \multicolumn{2}{c}{Agreement Rate (\%)} \\
\cmidrule(l){4-5}
Country & Official Name & Position (Abbreviated) & Human & Exa \\
\midrule
CZE & Pavel Blazek & Min. of Justice & 93.5 & 98.1 \\
CZE & Jaromir Drabek & Min. of Labor \& Social Affairs & 89.0 & 98.8 \\
JPN & Aiko Shimajiri & Min. in Charge of ``Cool Japan'' Strategy & 91.4 & 99.6 \\
JPN & Kenichiro Sasae & Ambassador to the US & 95.0 & 97.7 \\
SVK & Frantisek Ruzicka & Permanent Rep. to the UN (NY) & 92.1 & 99.4 \\
SVK & Pavol Pavlis & Min. of Economy & 92.4 & 98.4 \\
DNK & Rasmus Prehn & Min. for Development Cooperation & 94.9 & 99.7 \\
DNK & Kirsten Brosbol & Min. of Environment & 90.0 & 97.9 \\
KOR & Ju Chul-Ki & Senior Sec. for Foreign Affairs \& Security & 88.3 & 98.4 \\
KOR & Lee Byung-Ho & Dir., National Intelligence Service & 90.5 & 99.6 \\
COL & Alfonso Gomez Mendez & Min. of Justice \& Law & 88.2 & 98.0 \\
COL & Luis Felipe Henao Cardona & Min. of Housing \& Territorial Dev. & 91.6 & 97.7 \\
FIN & Jan Vapaavuori & Min. of Economic Affairs & 89.8 & 99.2 \\
FIN & Jari Lindstrom & Min. of Justice \& Employment & 94.0 & 99.3 \\
SVN & Bostjan Zeks & Min. w/o Portfolio (Slovenians Abroad) & 92.0 & 98.9 \\
SVN & Gorazd Zmavc & Min. w/o Portfolio (Slovenians Abroad) & 91.2 & 98.8 \\
IRL & Anne Colette Anderson & Permanent Rep. to the UN (NY) & 90.7 & 99.7 \\
IRL & Katherine Zappone & Min. for Children \& Youth Affairs & 88.4 & 97.8 \\
\bottomrule
\multicolumn{5}{l}{\footnotesize \textit{Note:} Agreement Rate indicates the percentage of claims where the auditor (Human or Exa) reached the} \\
\multicolumn{5}{l}{\footnotesize same verification verdict (Supported/Unsupported) as the automated CGT pipeline.} \\
\end{tabular}
\end{table}

\newpage
\section{Supplementary Results}
\label{sec:appendix_supplementary_results}

This section presents additional figures and analyses referenced in the main text to support key empirical claims. We focus on: (1) model performance without external resources, (2) comparison of agent-synthesized and Wiki-based corpora, (3) diagnostic checks of model heterogeneity under long-context constraints, (4) corpus composition and compression, (5) cross-national heterogeneity in retrieved corpora composition, and (6) granular mechanism plots illustrating recall dynamics and language effects.

\subsection{Model performance without external resources}
\begin{figure}[htbp]
  \centering
  \includegraphics[width=0.9\textwidth]{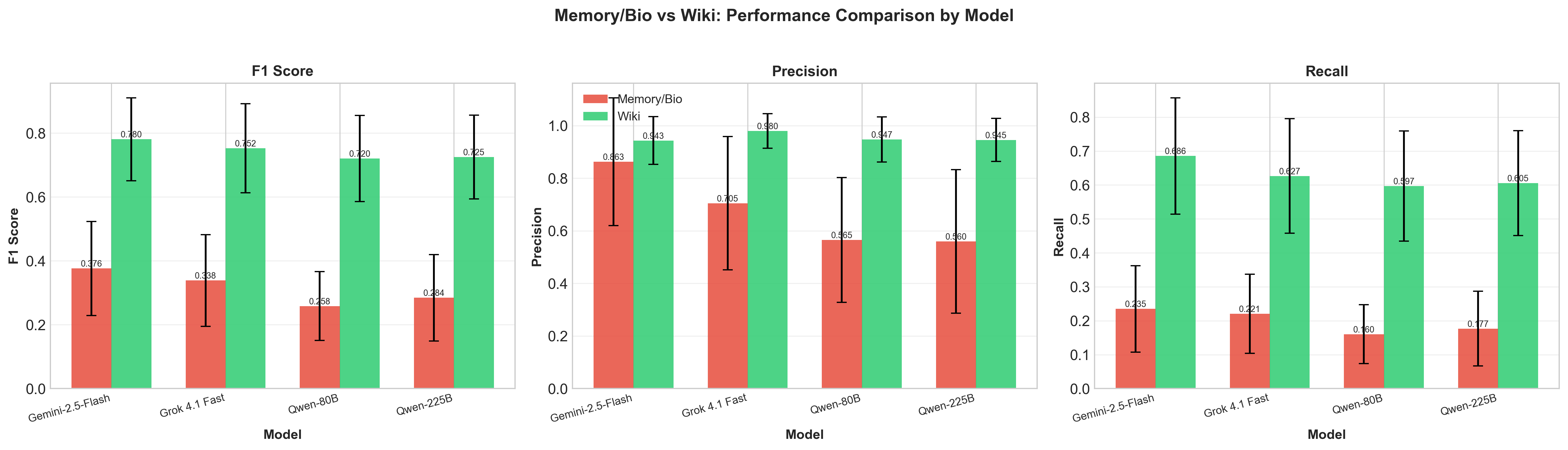}
  \caption{Three-panel comparison showing poor model performance without external resources on both precision and recall. Models with access to external resources (web search, retrieved documents) significantly outperform those operating without such access.}
  \label{fig:three_panel_comparison}
\end{figure}

Figure~\ref{fig:three_panel_comparison} illustrates that models exhibit poor performance on both precision and recall when operating without external resource access (e.g., without web search or retrieved documents). This three-panel comparison highlights the substantial performance gap between models with and without access to external information sources.

\subsection{Agentic Synthesis in a High-Curation Setting: China}
\label{sec:appendix_china_synthesis} 

\begin{figure}[htbp]
  \centering
  \includegraphics[width=0.95\textwidth]{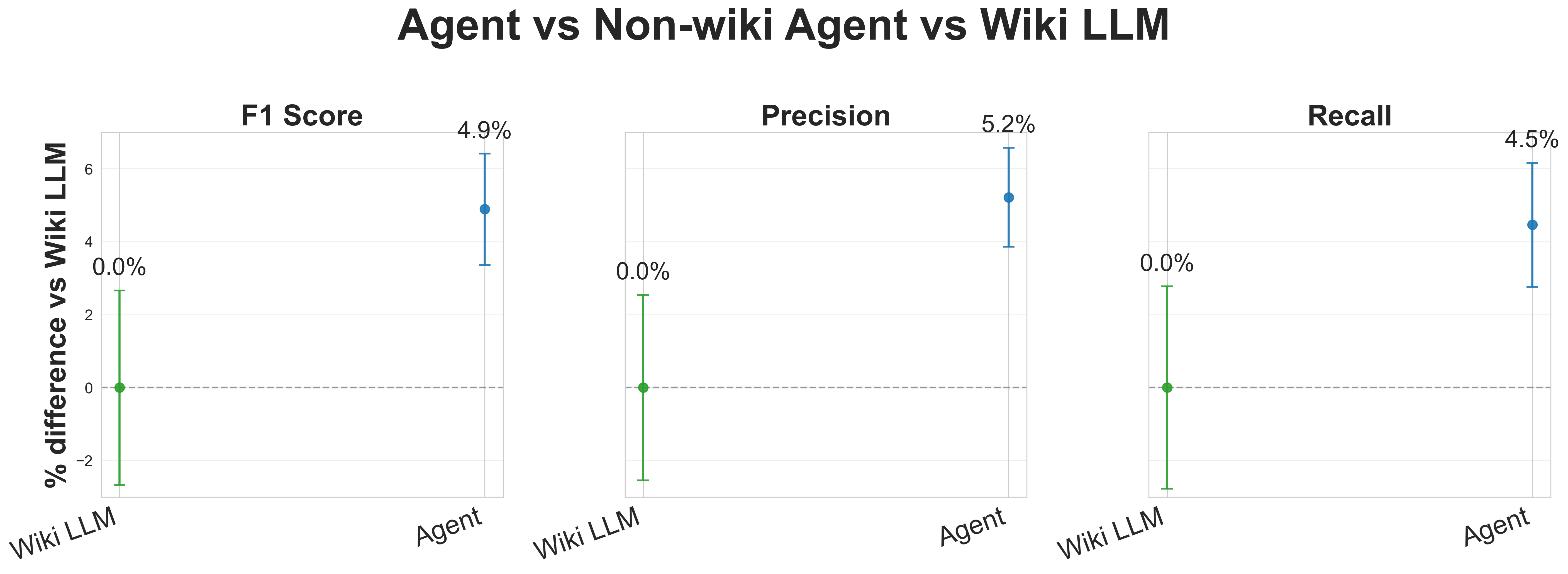}
  \caption{China setting: Comparison of agent-synthesized biographies and encyclopedic long-context baseline.}
  \label{fig:agent_vs_wiki_china}
\end{figure}

As a supplementary analysis, we examine agentic synthesis performance in the China setting, where the encyclopedic baseline is unusually strong. Unlike the U.S. and OECD samples discussed in the main text, Chinese political elites benefit from highly standardized and centrally curated biographical documentation, and Baidu Baike entries are typically comprehensive. As a result, any gains from synthesis are expected to be modest if the agent primarily reproduces already well-documented information. Figure~\ref{fig:agent_vs_wiki_china} reports the comparison between agent-synthesized biographies and the Wiki long-context baseline in this setting. Consistent with expectations, the magnitude of improvements is substantially smaller than in the U.S. and OECD samples. The agent increases F1 by 2.7 percentage points and recall by 3.8 points, accompanied by a modest precision gain of 1.4 points.

These results serve two purposes. First, they confirm that agentic synthesis does not degrade performance in environments where curated encyclopedic coverage is already strong. Second, the presence of small but detectable recall gains indicates that even in highly curated contexts, synthesis can recover incremental information omitted from baseline entries, such as minor concurrent appointments or short transitional roles. Together, the China results reinforce the interpretation of the main findings: the value of agentic synthesis scales with gaps in existing curation, yielding large gains where coverage is incomplete and converging toward parity where high-quality encyclopedic resources already exist.

\subsection{Model heterogeneity under long-context conditions}
\begin{figure}[htbp]
  \centering
  \includegraphics[width=0.9\textwidth]{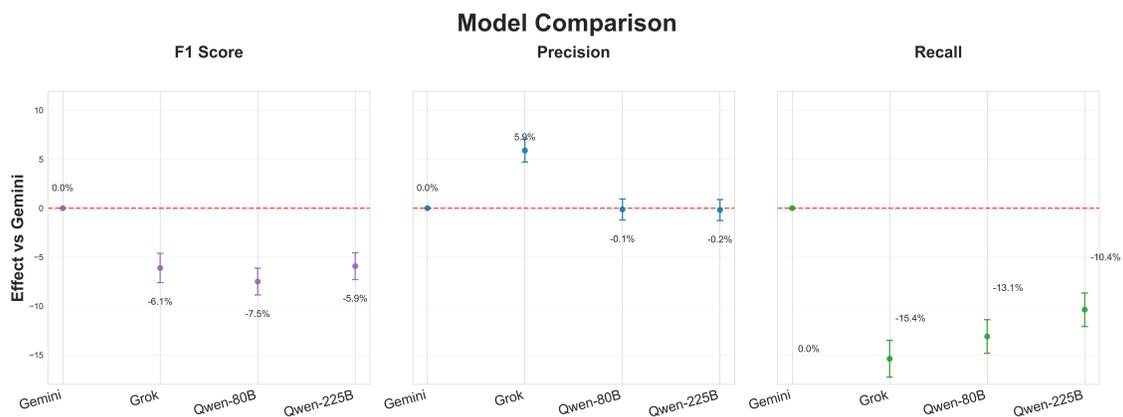}
  \caption{Performance differences across coder models (experiment-controlled) under long-context conditions, including both \texttt{LLM\_raw} and \texttt{LLM\_refined} variants.}
  \label{fig:model_comparison_experiment_controlled}
\end{figure}

We visualize performance disparities between coder models under uniform, long-context scenarios. Figure~\ref{fig:model_comparison_experiment_controlled} documents how both raw and refined pipelines (\texttt{LLM\_raw} and \texttt{LLM\_refined}) are affected.

\begin{figure}[htbp]
  \centering
  \includegraphics[width=0.9\textwidth]{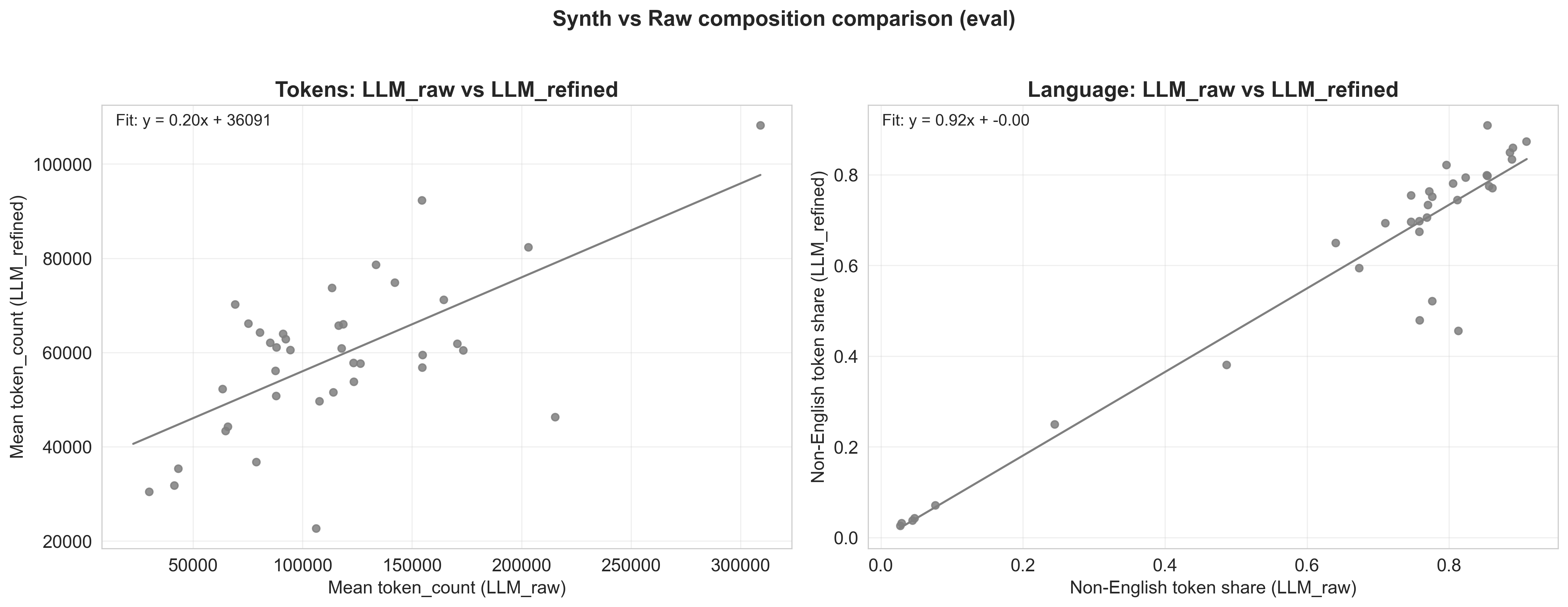}
  \caption{Token length and language composition for raw vs.\ refined corpora: refinement compresses token count while retaining high compositional correspondence.}
  \label{fig:composition_comparison}
\end{figure}

\subsection{Cross-national heterogeneity in retrieved corpora composition}

\begin{figure}[htbp]
  \centering
  \includegraphics[width=0.9\textwidth]{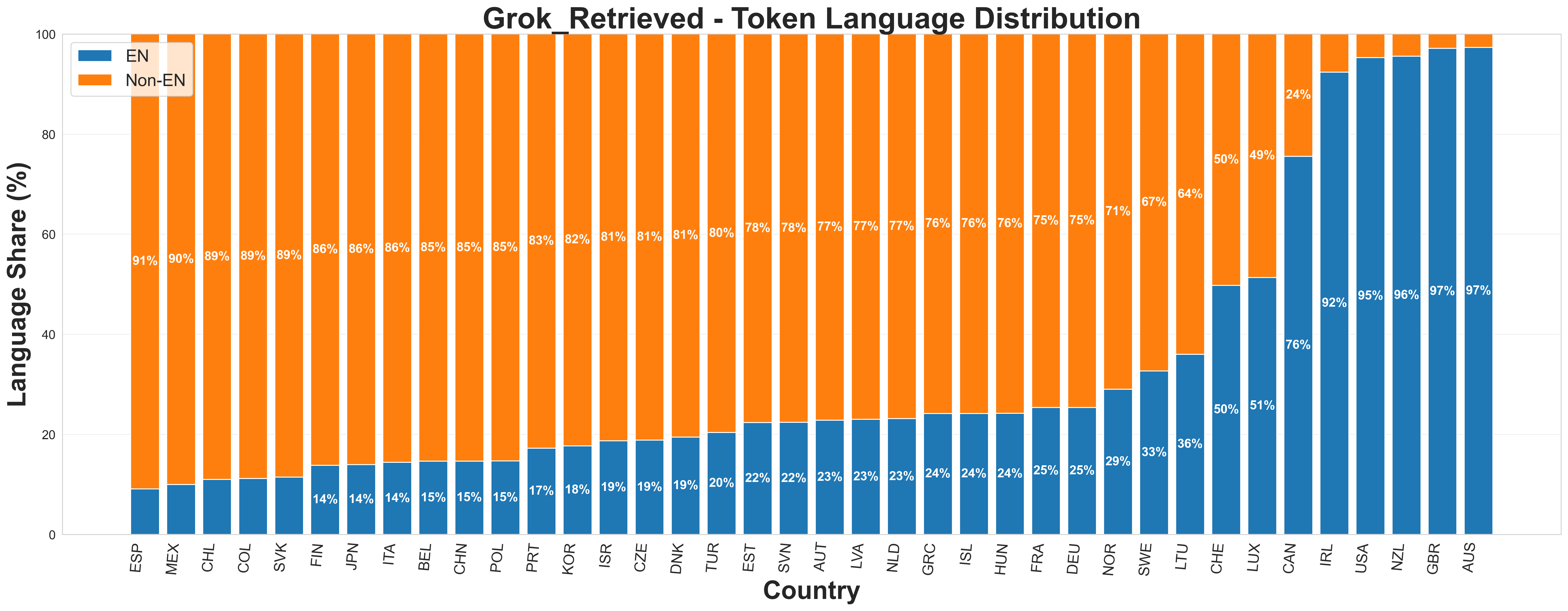}
  \caption{Cross-national heterogeneity in retrieved corpora language composition (descriptive analysis). Each bar represents the percentage of non-English resources in the retrieved corpus for each country, illustrating substantial variation in language diversity across the OECD sample. Higher non-English shares are associated with lower extraction performance.}
  \label{fig:language_descriptive}
\end{figure}

Figures~\ref{fig:composition_comparison}, and~\ref{fig:language_descriptive} illustrate cross-national heterogeneity in retrieved corpora composition. Figure~\ref{fig:composition_comparison} shows that refinement compresses token count while retaining high compositional correspondence between raw and refined corpora. Figure~\ref{fig:language_descriptive} shows the percentage of non-English resources in each country's retrieved corpus, illustrating substantial variation in language diversity across the OECD sample.

\subsection{Disaggregated mechanism plots}
\begin{figure}[htbp]
  \centering
  \includegraphics[width=0.9\textwidth]{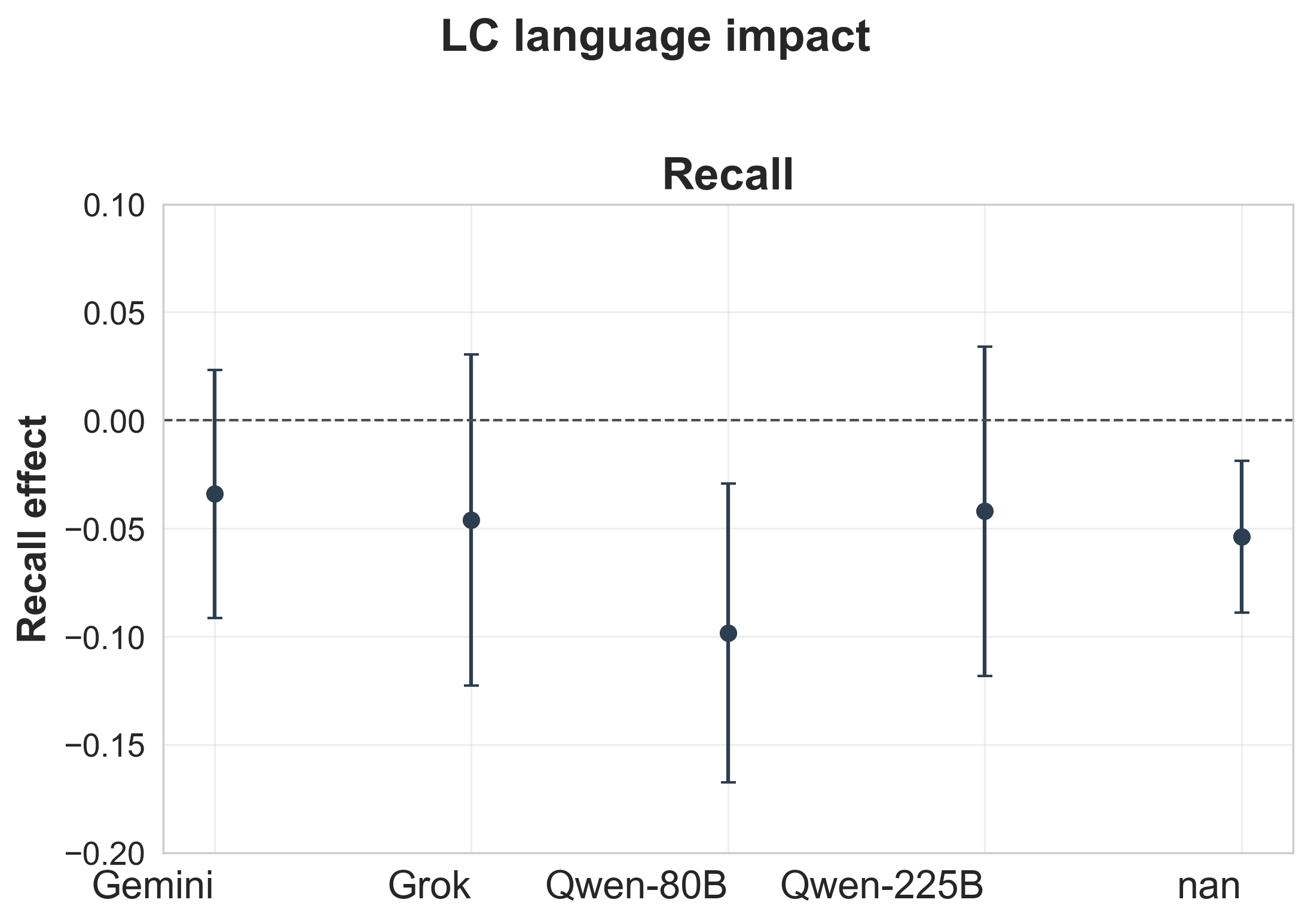}
  \caption{Language composition effects on recall (country fixed effects). Higher non-English token shares are associated with lower recall, indicating a quality channel in cross-national retrieval.}
  \label{fig:language_penalty}
\end{figure}

\begin{figure}[htbp]
  \centering
  \includegraphics[width=0.9\textwidth]{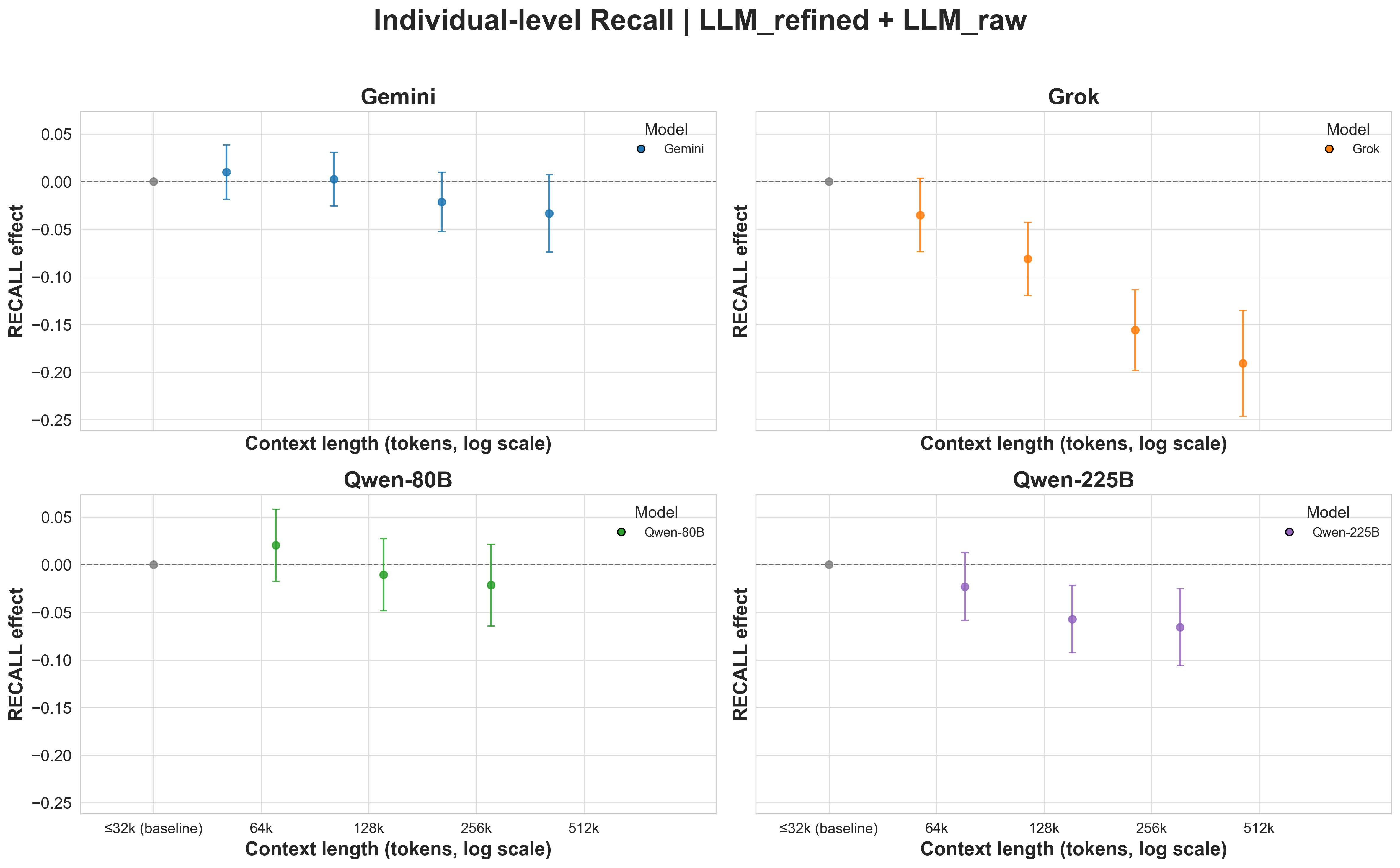}
  \caption{Binned recall by context length, disaggregated by model.}
  \label{fig:token_decay_by_model}
\end{figure}

 The following plots provide granular diagnostics on recall as a function of context length and language. 
Figure~\ref{fig:language_penalty} shows the language composition effects on recall with country fixed effects, demonstrating that higher non-English token shares are associated with lower recall.  Figure \ref{fig:token_decay_by_model} shows the token length effects on recall accross models , demonstrating that higher token length are associated with lower recall.

\section{Case Study: Erik Solheim Agent Run}
\label{sec:appendix_case_study_full}

This case study details the complete agent execution process for retrieving and synthesizing information about Erik Solheim, former Norwegian Minister of the Environment (2007--2012) and Executive Director of UN Environment Programme (2016--2018). We illustrate how the Supervisor--Searcher architecture (Section~\ref{sec:appendix_architecture}) operates in practice for a Non-US political figure requiring multilingual evidence synthesis.

\subsection{Agent Execution Overview}

The agent processed this case through \textbf{31 API calls} across 3 systematic batches, conducting \textbf{12 web searches} that returned 156 results, from which 14 documents were retrieved and archived. Table~\ref{tab:solheim_metrics} summarizes the key metrics.

\begin{table}[h]
\centering
\small
\caption{Erik Solheim Agent Run: Key Metrics}
\label{tab:solheim_metrics}
\begin{tabular}{@{}lr@{}}
\toprule
\textbf{Metric} & \textbf{Value} \\
\midrule
Total API Calls & 31 messages \\
Total Input Tokens & 474,595 \\
Total Output Tokens & 35,583 \\
Total Tokens Used & 510,178 \\
Web Searches & 12 searches (9 rounds) \\
Search Results Returned & 156 results \\
URLs Retrieved & 14 documents \\
Retrieved Chunks & 18 chunks \\
Processing Batches & 3 batches \\
Final Report Length & 4,371 characters \\
\bottomrule
\end{tabular}
\end{table}

\subsection{Three-Phase Search Strategy}

The agent execution followed an iterative refinement pattern, transitioning from broad information gathering to targeted gap-filling:

\subsubsection{Phase 1: Initial Skeleton Construction (Messages 0--7)}

\textbf{Supervisor's Goal:} Create a comprehensive initial sweep prioritizing official Norwegian government sources and Wikipedia to establish baseline biographical details.

\textbf{Searcher Execution:}
\begin{itemize}
    \item \textbf{Query:} \texttt{Erik Solheim AND (biografi OR miljøminister OR SV OR født)}
    \item \textbf{Target Sites:} \texttt{no.wikipedia.org}, \texttt{regjeringen.no}, \texttt{stortinget.no}
    \item \textbf{Language:} Norwegian (NO) and English (EN)
\end{itemize}

\textbf{Key Evidence Retrieved:}
\begin{itemize}
    \item \textit{no.wikipedia.org}: ``Erik Solheim (født 18. januar 1955)... Sjef for FNs miljøprogram... SV... MDG (2019--)''
    \item \textit{regjeringen.no (Government CV)}: Confirmed unique combined portfolio of Minister of the Environment and International Development; SV leadership 1987--1997
    \item \textit{stortinget.no (Parliament)}: ``Solheim, Erik (1955-)... Foreldre: Bjørn Elling Solheim og Sophie Grindstad... Cand.mag. 1980''
\end{itemize}

\textbf{Archive Update:} Identity confirmed (born 1955); baseline timeline established for SV leadership (1987--97), ministerial roles (2005--12), and UNEP (2016--18). Gaps identified: spouse names, early schooling details, and 1985--1987 career gap.

\subsubsection{Phase 2: Gap Filling -- Family and Mid-Career (Messages 8--20)}

\textbf{Supervisor's Goal:} Target Phase 2 expansion for precise details on early education, spouse names, career gaps (specifically 1985--1987 at Norges Handikapforbund), and post-2018 positions.

\textbf{Searcher Execution:}
\begin{itemize}
    \item \textbf{Query 1:} \texttt{Erik Solheim AND (kone OR Sissel Bugge OR Gry Ulverud)}
    \item \textbf{Query 2:} \texttt{Erik Solheim AND (Norges Handikapforbund OR konsulent)}
    \item \textbf{Query 3:} \texttt{Erik Solheim AND (post-2018 OR MDG OR Green Belt)}
\end{itemize}

\textbf{Key Evidence Retrieved:}
\begin{itemize}
    \item \textit{snl.no} (Store norske leksikon): ``Gift 1) 1989 med Sissel Bugge... Gift 2) 2004 med Gry Høeg Ulverud... Konsulent i Norges Handikapforbund 1985--1987''
    \item \textit{linkedin.com}: Post-2018 roles including ``President Europe-Asia Center (Jul 2023--)'' and ``Champion Global Renewables Alliance (Nov 2024--)''
\end{itemize}

\textbf{Archive Update:} Validated marriages (Sissel Bugge 1989--1998, Gry Ulverud 2004--present) and children; confirmed Consultant role at Norges Handikapforbund (1985--87); mapped extensive post-2018 advisory roles.

\subsubsection{Phase 3: Deep Dive -- Early Life and Education (Messages 21--40)}

\textbf{Supervisor's Goal:} Identify primary and secondary schools attended before Oslo katedralskole; check memoirs or interviews for early life details in the Bislett area.

\textbf{Searcher Execution:}
\begin{itemize}
    \item \textbf{Query 1:} \texttt{Erik Solheim AND (skole OR Bolteløkka OR Ila OR Bislett)}
    \item \textbf{Query 2:} \texttt{Erik Solheim AND (oppvekst OR barndom OR tidlig liv)}
    \item \textbf{Language:} Norwegian (NO) only
\end{itemize}

\textbf{Key Evidence Retrieved:}
\begin{itemize}
    \item \textit{vg.no (Factbox)}: ``...Oppvokst på Bislett, gikk på Bolteløkka skole, Ila skole og Oslo katedralskole''
    \item \textit{nettavisen.no}: ``...Vokst opp: Bislett... Bor nå: St. Hanshaugen''
\end{itemize}

\textbf{Archive Update:} Success (``needle in the haystack''); identified primary school \textit{Bolteløkka skole} and lower secondary \textit{Ila skole}. Timeline now continuous from childhood through 2025.

\subsection{Source Diversity and Language Composition}

The agent successfully navigated multilingual evidence, retrieving documents across multiple source types:

\begin{table}[h]
\centering
\small
\caption{Source Distribution for Erik Solheim Retrieval}
\label{tab:solheim_sources}
\begin{tabular}{@{}lcl@{}}
\toprule
\textbf{Source Type} & \textbf{Count} & \textbf{Examples} \\
\midrule
Government/Official & 6 & regjeringen.no CVs, stortinget.no \\
Reference/Encyclopedia & 2 & no.wikipedia.org, snl.no \\
News Media & 5 & vg.no, nettavisen.no \\
Professional & 2 & linkedin.com, hydropower.org \\
Other & 1 & oslobyleksikon.no, geni.com \\
\bottomrule
\end{tabular}
\end{table}

The search strategy demonstrates \textbf{adaptive multilingual retrieval}: initial queries combined Norwegian terms (``født'', ``kone'', ``skole'') with English disambiguation, prioritizing high-credibility Norwegian government sources while using English for cross-verification. This reflects the language composition patterns shown in Figure~\ref{fig:language_descriptive}.

\subsection{Retrieved Evidence and Archive Structure}

The agent archived 18 content chunks across 14 documents. Table~\ref{tab:solheim_archive} shows the archived entries with their evidence support categories under the FactNet protocol.

\begin{table}[h]
\centering
\scriptsize
\setlength{\tabcolsep}{2pt}
\caption{Case Study (Grok Candidates): Archived Entries and Evidence Support}
\label{tab:solheim_archive}
\begin{tabular}{@{}p{0.10\linewidth}p{0.74\linewidth}p{0.12\linewidth}@{}}
\toprule
\textbf{Type} & \textbf{Candidate Entry} & \textbf{Support} \\
\midrule
Education & 1961.01--1969.12 \textbar{} Bolteløkka skole \textbar{} Primary school & FULLY\_SUPPORTED \\
Education & 1969.01--1972.12 \textbar{} Ila skole \textbar{} Lower secondary & FULLY\_SUPPORTED \\
Education & 1970.01--1974.12 \textbar{} Oslo katedralskole \textbar{} High school & FULLY\_SUPPORTED \\
Education & 1974.01--1980.12 \textbar{} Universitetet i Oslo (UiO) \textbar{} Master & FULLY\_SUPPORTED \\
Party & 1977.01--1997.05 \textbar{} Sosialistisk Venstreparti (SV) \textbar{} Member/Leader & FULLY\_SUPPORTED \\
Party & 1977.01--1980.12 \textbar{} Sosialistisk Ungdom (SU) \textbar{} Leader & FULLY\_SUPPORTED \\
Party & 1981.01--1985.12 \textbar{} Sosialistisk Venstreparti (SV) \textbar{} Partisekretær & FULLY\_SUPPORTED \\
Party & 1987.01--1997.05 \textbar{} Sosialistisk Venstreparti (SV) \textbar{} Party Leader & FULLY\_SUPPORTED \\
Party & 2019.01--Present \textbar{} Miljøpartiet De Grønne (MDG) \textbar{} Member/Advisor & FULLY\_SUPPORTED \\
Career & 1985.01--1987.12 \textbar{} Norges Handikapforbund \textbar{} Konsulent & FULLY\_SUPPORTED \\
Career & 1989.10--1993.09 \textbar{} Stortinget \textbar{} Stortingsrepresentant Sør-Trøndelag & FULLY\_SUPPORTED \\
Career & 1993.10--2001.09 \textbar{} Stortinget \textbar{} Stortingsrepresentant Oslo & FULLY\_SUPPORTED \\
Career & 2000.03--2005.12 \textbar{} Utenriksdepartementet (UD) \textbar{} Spesialrådgiver & FULLY\_SUPPORTED \\
Career & 2005.10--2007.10 \textbar{} Utenriksdepartementet (UD) \textbar{} Utviklingsminister & FULLY\_SUPPORTED \\
Career & 2007.10--2012.03 \textbar{} Miljøverndepartementet / UD \textbar{} Miljøvernminister + Utviklingsminister & FULLY\_SUPPORTED \\
Career & 2013.01--2016.12 \textbar{} OECD \textbar{} Chair DAC & FULLY\_SUPPORTED \\
Career & 2016.01--2018.11 \textbar{} UN Environment Programme \textbar{} Executive Director & FULLY\_SUPPORTED \\
Career & 2017.01--Present \textbar{} BRIGC / Green Belt and Road Institute \textbar{} President & FULLY\_SUPPORTED \\
Career & 2019.01--2023.12 \textbar{} APRIL / WRI / TREELION \textbar{} Advisor & FULLY\_SUPPORTED \\
Career & 2023.07--Present \textbar{} Europe-Asia Center \textbar{} President & FULLY\_SUPPORTED \\
Career & 2024.11--Present \textbar{} Global Renewables Alliance \textbar{} Champion & FULLY\_SUPPORTED \\
Relatives & father & FULLY\_SUPPORTED \\
Relatives & mother & FULLY\_SUPPORTED \\
Relatives & ex-spouse & FULLY\_SUPPORTED \\
Relatives & spouse & FULLY\_SUPPORTED \\
Relatives & child & FULLY\_SUPPORTED \\
\bottomrule
\end{tabular}
\end{table}

\subsection{Ground Truth Comparison}

Table~\ref{tab:solheim_cgt} presents the consolidated ground truth (CGT) biography entries and their match categories against the agent output. This comparison reveals both the strengths and limitations of the agentic retrieval process.

\begin{table}[h]
\centering
\scriptsize
\setlength{\tabcolsep}{2pt}
\caption{Case Study (Ground Truth): CGT Entries and Match Categories}
\label{tab:solheim_cgt}
\begin{tabular}{@{}p{0.10\linewidth}p{0.74\linewidth}p{0.12\linewidth}@{}}
\toprule
\textbf{Type} & \textbf{CGT Entry} & \textbf{Match} \\
\midrule
Education & 1961.01--1969.12 \textbar{} Bolteløkka skole \textbar{} Primary school & FULL\_MATCH \\
Education & 1969.01--1972.12 \textbar{} Ila skole \textbar{} Lower secondary & FULL\_MATCH \\
Education & NA--1974.01 \textbar{} Oslo Cathedral School \textbar{} High school & FULL\_MATCH \\
Education & 1975.01--1980.01 \textbar{} University of Oslo \textbar{} cand.mag. & FULL\_MATCH \\
Party & 1977.01--1981.01 \textbar{} Socialist Youth \textbar{} Leader & FULL\_MATCH \\
Party & 1981.01--1985.01 \textbar{} Socialist Left Party \textbar{} Party Secretary & FULL\_MATCH \\
Party & 1985.01--1987.12 \textbar{} Socialist Left Party \textbar{} Central Exec. & NO\_MATCH \\
Party & 1987.04--1997.05 \textbar{} Socialist Left Party \textbar{} Party Leader & PARTIAL\_MATCH \\
Party & 1989.10--2019.01 \textbar{} Socialist Left Party \textbar{} Member & PARTIAL\_MATCH \\
Party & 2019.01--Present \textbar{} Green Party \textbar{} Member & FULL\_MATCH \\
Career & 1974.01--1975.01 \textbar{} Norwegian Air Force \textbar{} Conscript & NO\_MATCH \\
Career & 1985.01--1987.12 \textbar{} Norges Handikapforbund \textbar{} Consultant & FULL\_MATCH \\
Career & 1989.10--2001.09 \textbar{} Parliament of Norway \textbar{} Member of Parliament & FULL\_MATCH \\
Career & 2000.03--2005.10 \textbar{} Ministry of Foreign Affairs \textbar{} Special Adviser & FULL\_MATCH \\
Career & 2005.10--2012.03 \textbar{} Government of Norway \textbar{} Minister of International Development & FULL\_MATCH \\
Career & 2007.10--2012.03 \textbar{} Government of Norway \textbar{} Minister of the Environment & FULL\_MATCH \\
Career & 2012.03--2013.01 \textbar{} Ministry of Foreign Affairs \textbar{} Special Adviser & NO\_MATCH \\
Career & 2013.01--2016.06 \textbar{} OECD \textbar{} Chair of DAC & FULL\_MATCH \\
Career & 2016.06--2018.11 \textbar{} UN Environment Programme \textbar{} Executive Director & PARTIAL\_MATCH \\
Career & 2018.11--Present \textbar{} Belt and Road Green Development Coalition \textbar{} Vice President & PARTIAL\_MATCH \\
Career & 2018.11--Present \textbar{} Climate Council of Chief Minister MK Stalin \textbar{} Member & NO\_MATCH \\
Career & 2018.11--Present \textbar{} Global Solar Council \textbar{} Global Ambassador & NO\_MATCH \\
Career & 2018.11--Present \textbar{} Global Wind Energy Council \textbar{} Adviser & NO\_MATCH \\
Career & 2018.11--Present \textbar{} Green Hydrogen Organization \textbar{} Chairman & PARTIAL\_MATCH \\
Career & 2018.11--Present \textbar{} International Hydropower Association \textbar{} Board Member & NO\_MATCH \\
Career & 2019--Present \textbar{} Green Belt and Road Institute \textbar{} President & FULL\_MATCH \\
Career & 2019--Present \textbar{} World Resources Institute \textbar{} Senior Adviser & FULL\_MATCH \\
Career & 2019.05--Present \textbar{} Plastic REVolution Foundation \textbar{} CEO & NO\_MATCH \\
Relatives & father & FULL\_MATCH \\
Relatives & mother & FULL\_MATCH \\
Relatives & former spouse & FULL\_MATCH \\
Relatives & spouse & FULL\_MATCH \\
Relatives & child & FULL\_MATCH \\
\bottomrule
\end{tabular}
\end{table}

\subsection{Key Insights and Analysis}

\subsubsection{Discovery Successes}

The agent demonstrated strong performance on several fronts:

\begin{enumerate}
    \item \textbf{Iterative Refinement:} Successfully transitioned from broad queries (``biografi OR miljøminister'') to targeted searches (``Bolteløkka OR Ila''), demonstrating adaptive query reformulation.

    \item \textbf{Long-Tail Recovery:} Recovered specific primary and secondary school names (Bolteløkka skole, Ila skole) that represent ``needle in the haystack'' information requiring precise Norwegian-language queries.

    \item \textbf{Cross-Source Synthesis:} Integrated information across Wikipedia, government CVs, parliamentary records, encyclopedia entries, and contemporary news sources to build a comprehensive timeline.

    \item \textbf{Recent Activity Tracking:} Successfully identified post-2018 positions including 2024 appointments (Global Renewables Alliance) through LinkedIn and news sources.
\end{enumerate}

\subsubsection{Coverage Limitations}

The comparison with CGT reveals systematic gaps:

\begin{enumerate}
    \item \textbf{Weakly Connected Nodes:} Several concurrent advisory roles (Global Solar Council, Global Wind Energy Council, International Hydropower Association) were missed, suggesting the agent did not exhaustively traverse all post-2018 organizational affiliations.

    \item \textbf{Minor Positions:} Shorter-term roles (Norwegian Air Force conscript 1974--75, Special Adviser 2012--13, Plastic REVolution Foundation CEO) were not discovered, indicating challenges with brief or less-documented career phases.

    \item \textbf{Granularity Gaps:} Party membership continuity (1989--2019) was captured as a consolidated period rather than the granular breakdown in CGT, reflecting codebook representation choices.
\end{enumerate}

\subsubsection{Efficiency Analysis}

Token usage breakdown reveals the cost structure of agentic synthesis:

\begin{table}[h]
\centering
\small
\caption{Token Usage Breakdown by Component}
\label{tab:solheim_tokens}
\begin{tabular}{@{}lrrr@{}}
\toprule
\textbf{Component} & \textbf{Input} & \textbf{Output} & \textbf{Total} \\
\midrule
Searcher Agent & 421,483 & 32,723 & 454,206 \\
Coder Agent & 53,112 & 2,860 & 55,972 \\
\midrule
\textbf{Total} & \textbf{474,595} & \textbf{35,583} & \textbf{510,178} \\
\bottomrule
\end{tabular}
\end{table}

The Searcher consumed 89\% of total tokens, reflecting the computational cost of processing retrieved documents. The average of 14,534 input tokens per searcher call indicates substantial context accumulation across the multi-turn conversation.

\section{Prompts}
\label{sec:appendix_prompts}

We list the exact prompt templates used in our pipeline, grouped by stage.

\renewcommand{\arraystretch}{1.05}
\scriptsize
\setlength{\tabcolsep}{3pt}
\begin{longtable}{@{}p{0.26\linewidth}p{0.74\linewidth}@{}}
\caption{Prompt templates used in the research.}\\
\toprule
\textbf{Stage} & \textbf{Prompt} \\
\midrule
\endfirsthead
\toprule
\textbf{Stage} & \textbf{Prompt} \\
\midrule
\endhead
Architecture & Supervisor prompt \\
Architecture & Searcher prompt (Archive on) \\
\midrule
Experiment & Query template (EN) \\
Experiment & Research plan template (EN) \\
\midrule
Evaluation & Fact-checking (related-content judge) prompt \\
Evaluation & Entrywise evaluation prompt \\
\bottomrule
\end{longtable}
\normalsize
\renewcommand{\arraystretch}{1.0}

\subsection{Architecture prompts}
\subsubsection{Supervisor prompt}
\begin{lstlisting}
You are the Supervisor for a multi-step deep web research agent.

You reason based on the structured state:
- Research request (user query, constraints, codebook)
- Search batch history (each batch_overview with supervisor_task_instruction, research_summary, detailed_analysis)
- todo_list (remaining search gaps with [k] counters)
- global_summary (running summary of findings so far)

Each turn you must:
1) Update `global_summary` so it is a readable, self-contained summary of all solid facts found so far.
2) Update `todo_list` so it reflects the remaining important gaps.
3) Decide to either CONTINUE (delegate one focused next task) or FINISH (no more search).

OUTPUT FORMAT (JSON ONLY, no extra text, no markdown fences):
{
  "todo_list": "...",
  "next_task_instruction": "... or null",
  "global_summary": "..."
}

Field rules:
- `global_summary`:
  - Treat as the single evolving research summary.
  - Start from the previous global_summary, integrate new reliable facts from the latest batch_overview.
  - Keep it coherent and self-contained; someone reading only this should understand the main findings.
- `todo_list`:
  - Text block listing remaining gaps, typically as lines like `[k] <gap description>` (plus optional headings).
  - When a gap is fully answered, remove it.
  - When partially answered, rewrite to express only what is still missing.
  - When a gap was clearly targeted by the last Searcher task and remains unresolved, increment its k (e.g. `[1]`->`[2]`->`[3]`).
  - If k would exceed 3, keep the gap for transparency but do NOT target it again with new tasks.
- `next_task_instruction`:
  - Non-empty string => CONTINUE mode.
  - null => FINISH mode.
  - Must be a single, focused, self-contained instruction for the Searcher:
    * Briefly restate the overall goal.
    * Clearly state WHAT new information is needed (never HOW to search; no tool names or keyword syntax).

CONTINUE mode (non-empty `next_task_instruction`):
- Use when there are still important gaps in todo_list that are plausibly answerable by web research (prefer k = 1 or 2).
- Decompose broad gaps into concrete questions when possible (e.g. "exact dates for role X" instead of "complete career history").
- Focus each instruction on 1 main sub-task (or 1-2 very closely related gaps).

FINISH mode (`next_task_instruction` = null):
- Use when remaining gaps are minor, low-value, or have high counters (>3), or the user's request is sufficiently answered.
- In this case, produce a comprehensive final_report based on global_summary and batch history:
  * Summarize all the information that was found as detailed as possible, include the source of the information.
  * Note any major remaining uncertainties or unsolved gaps.
  * Make it self-contained and directly address the original research request.

Today is {current_date}.
\end{lstlisting}

\subsubsection{Searcher prompt}
\begin{lstlisting}
You are a professional Search Agent executing a research task to search, browse, and retrieve as broad relevant information as possible. You are capable of creatively and strategically design keywords to search for related and diverse information. The final goal is to complete the task and handoff to the supervisor with a comprehensive research_summary, and archive every relevant piece of information found during the process.

### Understand the Task
- You receive a **self-contained task instruction** from the Supervisor that includes:
    - The overall research goal
    - A summary of what has been found so far
    - The specific objective for this search batch
    - Any relevant constraints
- Read the provided `current_task_instruction` carefully
- The instruction should contain all context you need (goal, prior findings, current objective)
- Focus on the **specific objective** stated in the instruction


## Your Core Action Loop
You search, retrieve, and archive to complete the task:
1. Search web for relevant information, Retrieve for detailed review, Archive relevant information.
2. Handoff to the supervisor if collected enough information.

### Execute Search
- Call `web_search(search_intent=...)` with a structured search plan
   - `any_of` means at least one of the terms in the list should appear in results.
   - `must_include` means all of the terms in the list must appear in results.
   - `must_not_include` means none of the terms in the list may appear in results.
   - Start broad, then narrow based on results
   - Adjust the terms in `must_include` and `any_of` to make the search more specific or more broad based on observed results.
   - Avoid overly restrictive `must_include` terms
   - Mention generic meta-words like biography, bio, profile in `any_of` instead of `must_include`
   - Only use site restrictions when REALLY necessary
   - Flexibly use keywords in different languages as appropriate
- You have *{max_search_attempts}* search attempts, use wisely.

### Retrieve URLs Content for Browsing
- After each `web_search` call, call `retrieve_documents(urls=[...])` for the **promising** URLs from the latest results.
- Select up to 10 promising URLs per retrieve call.
- Skip retrieving if no results appear relevant.

### Archive Relevant Documents
  - Archived information will be reviewed by the supervisor for reference  - For each relevant document found during browsing, call `archive_document(detailed_analysis=[...])`:
  - `url`: Document URL
  - `title`: Document title
  - `task_summary`: Summary of how this document addresses the task
  - `relevant_chunk_labels`: List of chunk labels for relevant paragraphs (e.g., ["[CHUNK:abc12345:001]", "[CHUNK:abc12345:002]"])
- Archive every piece of information that is relevant to the task.
- Should have archived all relevant documents by the time you handoff.

## Handoff to Supervisor
When the task is complete, call `handoff_to_supervisor_with_overview`:
- `research_summary`: Comprehensive narrative including:
  - **What was found**: Specific information with concrete details
  - **What is lacking**: Information not found or uncertain
- `search_intent_summary`: Feedback on search effectiveness:
  - `bad_must_include`: Terms that performed poorly
  - `good_any_of`: Terms that worked well
  - `search_languages`: Languages used in searches

## Tools (USE ONLY THESE)
- web_search(search_intent: object) - execute search
- retrieve_documents(urls: list[string]) - fetch and chunk document content from URLs
- archive_document(detailed_analysis: list[object]) - archive every relevant chunk found during browsing to storage for future reference
- handoff_to_supervisor_with_overview(research_summary: string, search_intent_summary: object) - final handoff

## Important:
# Maintain loops of search, retrieve, and archive to complete the task incrementally.
# Handoff when the task is complete.
# Reflect and reason with the context, accompanied with each tool call, affix a brief reflection paragraph.

## Context

Today is {current_date}.
\end{lstlisting}

\subsection{Experiment prompts}
\subsubsection{Query template (EN)}
\begin{lstlisting}
Find comprehensive public information about {current_name}, a political or public figure{country_clause}{occupation_clause}{year_clause}.

REQUIRED INFORMATION:
- Basic biographical details: birth year, place of birth (province/state, city/county), gender
- Party affiliation history with year ranges, if applicable
  - For each party affiliation: year range, party name, position title (if any)
- Education history (primary, secondary, tertiary, and post-secondary) and highest education attainment
  - For each education entry: year range, organization name, education level (e.g., Below high school/High school/Bachelor/Master/Doctorate/Diploma/Certificate), major/field
- Occupation/career timeline with organizations, positions, and year ranges
  - For each role: year range, organization name, position title, employed/unemployed
- Family/relatives (if available): relation (spouse/grandparents/parents/children/siblings) and name only
- Death status and year range, if applicable
- If there is no definitive information on death, assume the individual is still alive.

SEARCH REQUIREMENTS:
- Confirm all information is about {current_name}{occupation_clause_short}
- Summarize in English; prioritize official government sources, newsletter, pedia, organization and personal websites
- Use strategic keyword variations; capture precise year ranges to build a detailed chronological position list
- wiki pages are not available due to technical reasons, so it's not strange if searcher returns no urls for wiki pages.

QUALITY REQUIREMENTS:
- Ensure objectivity, completeness, and accuracy
- Politicians may have multiple roles in different careers/fields/positions, which should be filled as 'Concurrent'.
- Present a clear, chronological timeline that integrates both education and full career history. Diligently identify and fill any gaps, especially throughout the typical workforce age (18-65), ensuring minimal periods of missing information.
- Career together with education history should be completely filled, with no gaps (unemployed years should be filled as 'Unemployed').

OUTPUT FORMAT:
- Include a comprehensive narrative biography (>=600 characters) integrating all details.
- Include the source of the information, credible or not, ensure reproducibility.
\end{lstlisting}

\subsubsection{Research plan template (EN)}
\begin{lstlisting}
# Phase 1: Comprehensive Initial Sweep
1. Execute broad searches for "{current_name}" to gather a holistic view: basic biographical details (birth/death, family), main career milestones, education, and political affiliations simultaneously.
2. Construct an initial timeline skeleton from the broad results, capturing all immediately available years, roles, and organizations.
3. Identify unique identifiers (e.g., specific keywords, middle names, known associations) to disambiguate from homonyms.

# Phase 2: Targeted Expansion & Detail Enrichment
1. Leverage specific entities found in Phase 1 (e.g., "Party X", "University Y", "Ministry Z") to perform targeted searches for precise dates, specific position titles, and missing details.
2. Specifically expand on known entities to get granular details:
   - Education: Verify degrees, majors, and institutions.
   - Party History: Clarify roles and affiliation periods.
   - Career: Flesh out concurrent roles and specific job titles using organization-specific keywords.

# Phase 3: Gap Analysis & Narrative Synthesis
1. Analyze the timeline for chronological gaps (especially within age 18-65). Perform specific queries to fill these gaps (e.g., check for private sector work or unlisted periods).
2. Re-verify any ambiguous data points (e.g., relatives, death date if unclear) and finalize the dataset.
3. Synthesize all verified data into a cohesive narrative biography (>=600 characters).
\end{lstlisting}

\subsection{Evaluation prompts}
\subsubsection{Fact-checking (related-content judge) prompt}
\begin{lstlisting}
You are a careful fact-checking assistant.

Your task is to evaluate **one biographical fact** about a person using ONLY the
provided related content (snippets aggregated from multiple URLs).

Person identifier: {official_id}
Person name: {official_name}

Biographical fact to check:
```text
{entry}
```

Related content (this is your ONLY evidence source; do not use outside knowledge):
```text
{related_content}
```

Instructions:
- Decide whether the fact is fully supported, partially supported, unclear, or contradicted
  by the related content.
- Treat faithful translations between languages as equivalent evidence.

Output JSON with exactly these fields:
- entry_text: the original fact text (string)
- verdict: 1-5 from not true to totally true, no option as unsure
- rationale: 1-3 sentences explaining your verdict, citing key phrases from the content
  (but do NOT invent new facts).

Do NOT include any commentary outside the JSON object.
\end{lstlisting}

\subsubsection{Entrywise evaluation prompt}
\begin{lstlisting}
## Task: Entrywise Biography Evaluation

You are an expert evaluator of biographical data extraction quality. Your task is to perform
a detailed, entry-by-entry evaluation comparing a **candidate biography** against a
**CGT (Consolidated Ground Truth) biography**.

---

## Person Information

- **Official ID**: {official_id}
- **Official Name**: {official_name}
- **Experiment Type**: {experiment_type}

---

## Core Evaluation Principle: Content Accuracy Over Structure

This is the most important guiding principle:
- When there are structural differences (e.g., CGT has one merged entry vs candidate has
  multiple split entries), prioritize judging whether the **total information content** is
  equivalent.
- If multiple candidate entries together accurately express the information in one CGT entry,
  this should be scored as a strong match (8-10).
- Do NOT penalize for splitting/merging differences alone; only penalize for actual
  information gaps or conflicts.

---

## Scoring Rubric (1-5 Scale)

### For CGT Entry Evaluation (How well is each CGT fact captured by the candidate?)

| Score | Category | Description |
|-------|----------|-------------|
| **5** | FULL_MATCH | Perfect or near-perfect match. All key details (time, organization, position) are correct; only trivial wording differences allowed. |
| **4** | PARTIAL_MATCH | Good match with small gaps or simplifications (e.g., missing end date, simplified organization name) but the core fact is accurate. |
| **3** | PARTIAL_MATCH | Partial match. The same event is referenced but with significant gaps or minor errors. |
| **2** | WEAK_MATCH | Very weak/unclear match. Only loosely related content; most details are missing or wrong. |
| **1** | NO_MATCH | No match at all. The CGT fact is completely absent from the candidate biography. |

### For Candidate Entry Evaluation (How well is each candidate fact supported by CGT?)

| Score | Category | Description |
|-------|----------|-------------|
| **5** | FULLY_SUPPORTED | Fully or almost fully supported by CGT. Clear matching CGT entry with at most trivial differences. |
| **4** | PARTIALLY_SUPPORTED | Mostly supported. Core fact is in CGT, with small additions or wording differences. |
| **3** | PARTIALLY_SUPPORTED | Partially supported. Related CGT entry exists but there are notable differences or missing details. |
| **2** | WEAKLY_SUPPORTED | Weakly supported. Only loosely related CGT content; candidate may contain errors. |
| **1** | NOT_SUPPORTED | No support (hallucination). This candidate entry has no real basis in the CGT. |

---

## Difference Codes (for CGT evaluations with score < 5)

When a CGT entry is not perfectly matched, select applicable codes from:

| Code | Meaning |
|------|---------|
| `TIME_YEAR` | Year is incorrect |
| `TIME_MISSING` | Time information is missing from candidate |
| `ORG_WRONG` | Organization name is incorrect (not just abbreviation) |
| `POSITION_WRONG` | Position/title is incorrect |
| `POSITION_INCOMPLETE` | Missing concurrent positions or partial title |
| `EXTRA_INFO` | Candidate has extra information not in CGT (neutral/positive) |

---

## Flexible Alignment Rules

### 1-to-N Matching (CGT merged entry vs Candidate split entries)
- If the candidate splits one CGT entry into multiple lines, list all matching candidate
  entries separated by " || " in the `matched_candidate_entries` field.
- Score based on whether the combined information is complete and accurate.

### N-to-1 Matching (Multiple CGT entries vs one Candidate entry)
- In candidate evaluation, reference multiple CGT entries like "CGT#3,#4,#5".
- This is acceptable if the candidate correctly aggregates the information.

### Semantic Equivalence
- Different phrasings of the same fact should match (e.g., "Mayor" = "City Mayor").
- Abbreviations vs full names are acceptable (e.g., "EPA" = "Environmental Protection Agency").
- Cross-language translations are equivalent if semantically the same.

---

## Important Notes

1. **Be consistent**: Apply the same standards across all entries.
2. **Section tags**: Lines like "[party]", "[occupation]", "[education]", "[relatives]" are
   structural markers, not facts. Skip them when counting entries.
3. **Empty lines**: Ignore empty lines when counting and evaluating.
4. current date is 2025-11-25
---

## Input Data

### CGT BIOGRAPHY (Ground Truth):
```text
{cgt_biography}
```

### CANDIDATE BIOGRAPHY (Experiment: {experiment_type}):
```text
{experiment_biography}
```

---

## Output Format

Produce a JSON object with exactly these fields:

- `official_id`: string (copy from input: "{official_id}")
- `official_name`: string (copy from input: "{official_name}")
- `experiment_type`: string (copy from input: "{experiment_type}")
- `cgt_entry_count`: integer (number of non-empty, non-tag lines in CGT)
- `candidate_entry_count`: integer (number of non-empty, non-tag lines in candidate)
- `cgt_evaluations`: array of objects, one per CGT entry, each with:
  - `index`: integer (1-based)
  - `cgt_entry_text`: string (the CGT line)
  - `matched_candidate_entries`: string (matching candidate text or "NO_MATCH")
  - `match_score`: integer (1-5)
  - `match_category`: string ("FULL_MATCH", "PARTIAL_MATCH", "WEAK_MATCH", or "NO_MATCH")
  - `difference_codes`: array of strings (codes from the table above, or empty)
  - `notes`: string (brief explanation)
- `candidate_evaluations`: array of objects, one per candidate entry, each with:
  - `index`: integer (1-based)
  - `candidate_entry_text`: string (the candidate line)
  - `matched_cgt_entries`: string (e.g., "CGT#3" or "CGT#1,#2" or "NO_SUPPORT")
  - `support_score`: integer (1-5)
  - `support_category`: string ("FULLY_SUPPORTED", "PARTIALLY_SUPPORTED", "WEAKLY_SUPPORTED", or "NOT_SUPPORTED")
  - `notes`: string (brief explanation)
- `qualitative_summary`: string (2-4 sentences on overall quality)

Do not include markdown fences or any text outside the JSON object.
\end{lstlisting}

\section{A Practical Guide to Information Extraction with LLMs}
\label{sec:appendix_practical_guide}

This appendix provides a practical guide for applying Large Language Models (LLMs) to information extraction tasks in the social sciences. The guiding principle is to treat \textit{extraction} as an end-to-end data-production task and, when necessary, to separate it into two modular stages: \textbf{synthesis} (evidence acquisition and refinement) and \textbf{coding} (mapping a refined corpus into a structured codebook). This modular design improves auditability and helps us diagnose whether errors arise from missing evidence (a synthesis failure) or incorrect mapping (a coding failure).

\subsection{A minimal workflow for reliable extraction}
\paragraph{Step 0 (define the record and the evidence rule).}
Extraction is only well-defined relative to a codebook. Before using an LLM, we specify (i) a field-level codebook (variables, types, allowed formats), (ii) normalization rules (names, organizations, dates), and (iii) a groundedness requirement (what constitutes sufficient evidence for a claim). In political fact extraction, a small number of ambiguous fields (e.g., office titles, start/end dates) can drive large downstream measurement error, so explicit formatting and disambiguation rules are essential.

\paragraph{Step 1 (diagnose whether synthesis is necessary).}
The most important practical decision is whether the available sources are effectively \textbf{curated} or \textbf{open-ended and noisy}. When a short, high-signal source exists (e.g., Wikipedia, an official CV, a curated archive), we can often run \textbf{coding-only}: we provide the curated text and ask the model to output the structured record.\footnote{Even in curated settings, long contexts can degrade reliability when relevant facts are buried deep in lengthy inputs \parencite{liu2024lost}.} When relevant evidence is dispersed across many documents (e.g., the open web) or the total context exceeds any fixed window, \textbf{synthesis is necessary}: we must decide which sources to read and how to condense them into a signal-dense representation before coding can be valid.

\paragraph{Step 2 (implement coding with constraints and groundedness).}
The coding stage maps a fixed input corpus into a structured record. In practice, we recommend three safeguards: constrain outputs to be strictly codebook-conformant (e.g., JSON with fixed keys and date formats); require evidence pointers (quotes/snippets) for each claim to reduce hallucination risk \parencite{mallen2023not}; and, when feasible, separate generation from validation (a second pass or second model that checks codebook compliance and evidence support). Modern LLMs can often perform this stage in a zero-shot or few-shot manner when the input is curated and the codebook is explicit \parencite{ornstein2025train,ziems2024can}.

\paragraph{Step 3 (implement synthesis as bounded, credibility-aware refinement).}
Synthesis is an evidence-refinement process: retrieve, filter, cross-check, and compress information into a corpus that is feasible for coding. While there are multiple valid implementations (keyword search, embedding retrieval, human-in-the-loop), open-ended political fact extraction often requires an \textbf{interactive} process because early discoveries change what should be searched next. Agentic workflows operationalize this by alternating between reasoning and tool use (ReAct) \parencite{yao2023react}, enabling adaptive query refinement and iterative evidence accumulation. Operationally, we recommend explicit retrieval budgets (steps/tokens/sources), credibility-aware filtering (prioritize authoritative sources; deduplicate near-identical content), and compression with traceability (store a refined corpus plus source-linked snippets so claims remain auditable).

\paragraph{Step 4 (evaluate and interpret trade-offs).}
For extraction, both false positives and false negatives are substantively costly. Precision captures whether extracted claims are correct; recall captures whether the system recovers relevant claims; and F1 summarizes the trade-off. When precision is low, the coding stage is often hallucinating or mis-mapping (tighten groundedness and codebook constraints; improve synthesis filtering). When recall is low, the system is missing evidence (increase synthesis coverage or improve the refined representation).

\end{refsection}
\clearpage
\printbibliography

\end{document}